\documentclass[conference]{IEEEtran}
\usepackage{times}

\usepackage{multicol}
\usepackage[numbers]{natbib}
\usepackage{tikz} 
\usepackage{graphicx}
\graphicspath{{figures/}}
\usepackage{subfigure}

\begin{document}

\newcommand{\p}{\textbf{p}}
\newcommand{\ps}{\p^{\ast}}
\newcommand{\q}{\textbf{q}}
\newcommand{\qs}{\q^{\ast}}
\newcommand{\D}{\textbf{D}}
\newcommand{\A}{\textbf{A}}
\newcommand{\bb}{\textbf{b}}
\newcommand{\nred}[1]{{\vspace*{1em}\todo[inline,color=red!50]{#1}}}
\newcommand{\nblue}[1]{{\vspace*{1em}\todo[inline,color=blue!50]{#1}}}
\newcommand{\colr}[2]{\color{#2}#1\color{black}}
\newcommand{\red}[1]{\color{red}#1\color{black}}
\newcommand{\blue}[1]{\color{blue}#1\color{black}}
\newcommand{\aude}[1]{\color{green}#1\color{black}}
\newcommand{\unih}{\texttt{uni8}\ }
\newcommand{\uniD}{\texttt{uniD}\ }
\newcommand{\rdm}{\texttt{rdm}\ }

\usetikzlibrary{shadows,arrows}
% Define the layers to draw the diagram
\pgfdeclarelayer{background}
\pgfdeclarelayer{foreground}
\pgfsetlayers{background,main,foreground}
 
% Define block styles  
\tikzstyle{texto} = [text width=10em, text centered]
\tikzstyle{linepart} = [draw, thick, color=black!50, -latex', dashed]
\tikzstyle{line} = [draw, thick, color=black!50, -latex']
\tikzstyle{ur}=[draw, text centered, minimum height=0.01em]
 
% Define distances for bordering
\newcommand{\blockdist}{1.3}
\newcommand{\edgedist}{1.5}

% Draw background
\newcommand{\background}[5]{%
  \begin{pgfonlayer}{background}
    % Left-top corner of the background rectangle
    \path (#1.west |- #2.north)+(-0.5,0.7) node (a1) {};
    % Right-bottom corner of the background rectanle
    \path (#3.east |- #4.south)+(+0.5,-0.25) node (a2) {};
    % Draw the background
    \path[fill=gray!10,rounded corners, draw=black!50, dashed]
      (a1) rectangle (a2);
    \path (a1.east |- a1.north)+(+.8,-0.0) node (u1)[texto]
      {\scriptsize\textit{#5}};
  \end{pgfonlayer}}

\newcommand{\transreceptor}[3]{%
  \path [linepart] (#1.east) -- node [above]
    {\scriptsize Transreceptor #2} (#3);}

\tikzstyle{block_background}=[draw, fill=blue!20, text width=6.0em, text centered,
  minimum height=1.5em,drop shadow]
\tikzstyle{block} = [block_background, text width=8em, minimum width=10em,
  minimum height=3em, %rounded corners, 
  drop shadow]
\tikzstyle{data} = [text centered, text width=8em, minimum width=10em,
  minimum height=3em]
\newcommand{\block}[2]{node (p#1) [block]
  {#2}}
\newcommand{\data}[2]{node (p#1) [data]
  {#2}}

% paper title
\title{A Complete framework for ambush avoidance in realistic environments}

% You will get a Paper-ID when submitting a pdf file to the conference system
%\author{Author Names Omitted for Anonymous Review. Paper-ID [65]}

\author{\authorblockN{Emmanuel Boidot}
\authorblockA{
eboidot3@gatech.edu
%School of Aerospace Engineering\\
%Georgia Institute of Technology\\
%Atlanta, Georgia 30332--0250
}
\and
\authorblockN{Aude Marzuoli}
\authorblockA{
amarzuoli3@gatech.edu\\
School of Aerospace Engineering\\
Georgia Institute of Technology\\
Atlanta, Georgia 30332--0250
}
\and
\authorblockN{Eric Feron}
\authorblockA{
%School of Aerospace Engineering\\
%Georgia Institute of Technology\\
%Atlanta, Georgia 30332--0250
feron@gatech.edu}
}

% avoiding spaces at the end of the author lines is not a problem with
% conference papers because we don't use \thanks or \IEEEmembership
\vspace{-2em}

\maketitle

\begin{abstract}
Operating vehicles in adversarial environments between a recurring origin-destination pair requires new planning techniques. A Ruckle-inspired game is introduced. The goal of the first player is to minimize the expected casualties undergone by a convoy. The goal of the second player is to maximize this damage. The outcome of the game is obtained via a linear program that solves the corresponding minmax optimization problem over this outcome. 
%\aude{how?}
% construction
Different environment models are defined in order to compute routing strategies over unstructured environments. To compare these methods for increasingly accurate representations of the environment, a grid-based model is chosen to represent the environment and the existence of a sufficient network size is highlighted.
% framework
A global framework for the generation of realistic routing strategies between any two points is described. 
This framework requires a good assessment of the potential casualties at any location, therefore the most important parameters are identified. 
Finally the framework is tested on real world environments.% and the difference induced by players resources are illustrated. %\aude{and validated somehow?}.
\end{abstract}

\IEEEpeerreviewmaketitle

%%%%%%%%%%%%%%%%%%%%%%%%%%%%%%%%%%%%%%%%%%%%%%%%%%%%%%
\section{\label{sec:intro}Introduction}

%
%Context
%expliquer complexite du probleme, type de donnees, environnement 
%
%General Goal: route planning for vehicle in hostile environment
%
%Present Problem
%
%expliquer difference path/route
%
%ici questions fondamentales
%- nb de chemins caracteristiques
%- "entropy"
%
%what we do here is numerical investigation
%
%Related Work
%\nblue{-initial work: Ruckle
%
%Geometric games played on a finite lattice L will always have a solution because of the 
%Minimax Theorem of von Neumann \cite{vonNeumann-1947}}
%
%\nblue{-Base: Eric and Farmey
%
%mixed (stochastic/probabilistic) strategy as opposed to pure startegy.}
%
%\nblue{-Application: Swiss guys, piracy?}
%
%\nblue{-Dynamic: Emilio et Sertac \cite{DBLP:conf/wafr/KaramanF10}}
%
%\nblue{-Risk: Panos?}

%\nblue{Specific research question of this paper: 
%\red{0: risk related automated decision making against smart player }
%1. new methodology: exit the purely theoretical world of game theory in order to make realistic games simulation 
%2: while highlighting interesting results.}

We seek to develop routing strategies for dynamical systems operating in adversarial environments. For the problem at hand, an agent is trying to move a vehicle from a given origin a desired target set, while avoiding undesired set of states (ambushes), and another agent is rewarded if the system is caught in one of the ambushes. Examples of applications of this research could be found in protection of automated vehicles, such as delivery vehicles, or piracy prevention. 
%Given the recent development in automated systems and their increasing importance in our everyday life, the risk for potential attacks targeting these systems becomes concerning. A strategic routing strategy will most likely be of great use in a near future. The present paper tackles this problem from a game theoretic perspective with important insights from experiments.

Recently, significant research has focused on dynamic pursuit-evasion games \cite{DBLP:conf/wafr/KaramanF10}. %\red{description}. 
Unlike these efforts, our work is interested in computing mixed planning strategies. 
%While the pursuit-evasion games solvers return one course of action for a given scenario, our proposed approach produces a probabilistic distribution of trajectory. 
In the context of differential pursuit-evasion games, the solutions found are often found to be deterministic. Our problem returns a random distribution of solutions. This means that, for the same scenario, the returned strategy is unique but two vehicles following this strategy might use different paths. For clarity reasons, a probabilistic strategy (detailed later) will be described as a "route" and a deterministic realization of this strategy will be described  as a "path". 

The idea behind this research effort was first introduced by Ruckle \cite{ruckle1976ambushing}, who extended Isaacs's \cite{isaacs2012differential} classical battleship versus bomber duel by interpreting a two-dimensional environment as a rectangular array of lattice points. Ruckle stated that such geometric games played on a finite lattice always exhibit a solution because of the Minimax theorem of Von Newmann \cite{neumann44a}. He also identified necessary and sufficient conditions for the mixed strategy of a player to be optimal. Ruckle's idea was later extended by Farmey and Feron \cite{Farmey-thesis} in order to formulate this game as a two player zero sum  game applicable to a non-lattice network. %They proposed a linear optimization algorithm to solve the game. 
While they advanced the idea of a variable game outcome at each node of the network, no analysis of this parameter was performed. Their formulation was later used in applications for piracy prevention \cite{vanek-agent2013}\cite{vanek_327} and money transit protection \cite{salani2010ambush}, where the multiple destination case is tackled.

The specific goal in our approach is to exit the purely theoretical world of game theory in order to perform simulations close to real-life applications. This is achieved through two distinct processes. First, a method to create a network on any environment on which the game is to be run is developed. Second, the outcome of an ambush is no longer constant over the environment. The outcome of an ambush now depends on its position and the factors influencing this outcome are identified. Therefore the optimal routing strategy can be obtained on any environment without manual input of a case specific game matrix.
The authors believe that these two points are the main contribution of the present paper to the current state of the art in the field of ambush games.

The outline of this paper is as follows. In Section \ref{sec:approach} the game and its mathematical formulation are further described. Section \ref{sec:framework} details different network construction methods and ambush types considered for the adaptation of the game to unstructured environments, while Section \ref{sec:performance} discusses the results of the game for different methods and parameters. Section \ref{sec:applications} describes the factors of risk for ground vehicles and illustrates the use of our work on several applications.

%%%%%%%%%%%%%%%%%%%%%%%%%%%%%%%%%%%%%%%%%%%%%%%%%%%%%%
\section{\label{sec:approach}Approach}

	%%%%%%%%%%%%%%%%%%%%%%%%%%%%%%%%%%%%%%%%
	\subsection{Game Description}
	
%\nred{jaime pas utiliser le mot risk pour une notion qui represente des casualties -> "local outcome"?}
	
%\nblue{game}
The problem of interest is to plan a path for a convoy that needs to journey from an origin A to a destination B in an environment where hostile forces might try to ambush it. It is modelled as a two players non-cooperative zero-sum game, meaning that a player's sole goal is to optimize his personal gain (non cooperative) and the sum of the outcome over all players is zero. Player 1, BLUE, chooses a route from origin to destination. Player 2, RED, selects a number of locations at which to set up ambushes. The number of ambushes depends on RED's resources. In the remainder of this paper RED only owns one resource, i.e. it can only set one ambush. This assumption can be made without consequences because we are interested in mixed strategies. It makes the sum of probability over all possible states for RED equal to one instead of a higher integer.
If BLUE's path passes through an ambush site, then RED wins. If BLUE's path avoids all ambush sites, then BLUE wins. The outcome of an ambush corresponds to the casualties that the agent would experience in this specific area and it is dependent on the characteristics of the local environment.  The outcome of a game is the sum of this local outcome over all the ambushes that BLUE's path has gone through. 
The outcome $\alpha$ of an ambush at point $x$ satisfies $\alpha = f(x)$. It is assumed to be known by both players. Its approximation is discussed in Section \ref{sec:applications}.
%In the remainder of this proposal, this outcome will be referred to as the "risk" at this location. This notion has to be distinguished from the usual definition of risk, which would be the probability that RED sets an ambush at a given location.

%\nblue{Environment, Information sharing}

The environment is represented by a network $(\mathcal{N},\mathcal{E})$ and a local outcome map. Each ambush area is associated with a local outcome $\alpha_i$. The set $\alpha$ is the discretized local outcome map, which measures the possible losses for BLUE. In this paper only single-stage ambush games are considered, meaning that both players need to decide their strategy at the beginning of the game and cannot reconsider it during the game. Neither player has any information about the other player's current path. The environment and its discrete representation are the only information RED and BLUE share : the network, the local outcome map, BLUE's origin and destination points and the position of the ambush areas (but not the position of those chosen by RED) are known to both players.

%\nblue{original (Farmey)}	
		
A possible strategy for BLUE is represented by a probability vector $\p$, where $p_{ij}$ is the probability that the convoy uses edge $e_{ij}$ between nodes $n_i$ and $n_j$. Similarly, a strategy for RED is represented by a probability vector $\q$ that contains the probability $q_k$ that RED sets an ambush at the ambush area $a_k$. The physical meaning of an ambush area corresponds to the area RED can act on if he sets his ambush inside a given area. Setting an ambush on this area means that BLUE will be ambushed if its path goes through nodes in this set. 
%\aude{is an area defined by a set of edges or nodes?}.		
It is common for this type of game theoretic problems to assume that ambushes take place at nodes or on edges of the network, which is the approach adopted in previous work \cite{DBLP:conf/ssrr2012/Boidot}. However, to expand the approach to continuous state space, it makes more practical sense for RED's strategies to relate to areas of the environment (continuous sets). This way, RED's strategy translates into where it would position its ambushes given a specified action range. 

Given the network and the local outcome map, the goal of the game is to find the optimal strategy $\ps$ for BLUE, assuming that RED follows its optimal strategy $\qs$.
% mixed strat
Consider the case of a recurring continuous transition from state A to state B, where there exist two different paths from A to B. A deterministic approach will return only one of the two paths, whereas the mixed strategy $p$ will return a probability $p_1$ of using the first path and a probability $p_2$ of using the second path. Incorporating mixed strategy solutions means that, for a given environment, the opponent can at best (after a large number of runs following this strategy) figure out the strategy $p$ but will never gather any certitude over the convoy's exact path.

%\nblue{expansion (ambush area, network construction, etc)}

	%%%%%%%%%%%%%%%%%%%%%%%%%%%%%%%%%%%%%%%%
	\subsection{Mathematical Formulation}
%\nblue{original (Farmey)}

Let $(\mathcal{N},\mathcal{E})$ be a network. The most simple example case where ambushes happen at nodes is considered first. Therefore ambush areas $\{a_i\}$ and nodes $\{n_i\}$ represent the same set in this subsection. Let $\p$ (resp. $\q$) be the probability vector representative of BLUE's (resp. RED's) mixed strategy. Assume that the two players strategies are independent. At each node $n_j$, the probability that BLUE gets ambushed is equal to the probability that BLUE's path goes through $n_j$ times the probability that RED sets an ambush at this node. The gain for RED at this node being $\alpha_i$, the expected outcome of the game relative to this node is: $\sum\limits_{i | (i,j)\in \mathcal{E}} p_{ij} q_j \alpha_j$. Therefore the strategic outcome of the game is :
\begin{equation}
\mathcal{V} = \sum\limits_{j \in \mathcal{N}} \sum\limits_{i | (i,j)\in \mathcal{E}} p_{ij} q_j \alpha_j \\
		= \q^t \D \p. 
\label{eq:strategic_outcome}
\end{equation}
with $D_{jk} = \alpha_j$ if the $k^{th}$ line of \textbf{p} represents the probability that BLUE uses an edge $e_{ij}$ directed towards $n_j$, and $D_{jk} = 0$ otherwise.

The objective of the approach is to find the strategy for BLUE that minimizes the maximal possible outcome for RED.
%
%\begin{equation}
%\p^{\ast} = \arg\min\limits_{\p} \max\limits_{\q} \q^t \D \p. 
%\end{equation}
Provided that $q_j \leq 1\ \ \forall j$, RED can always maximize $\mathcal{V}$ by choosing the node $n_j$ for which the probability of BLUE passing through that node weighted by the value $\alpha_j$ is maximal. Therefore BLUE's optimal solution is to minimize this product across all nodes :

\begin{equation}
\p^{\ast} = \arg\min\limits_{\p} \left( \max\limits_{j \in \mathcal{N}} \sum\limits_{i | (i,j) \in \mathcal{E}} p_{ij} \alpha_j \right). 
\label{eq:p_opt_formulation}
\end{equation}

The other constraints of this problem enforce the flow conservation through the network. The probability that the convoy arrives at node $n_j$ is equal to the probability that the convoy leaves the same node. Probabilities of the convoy being at origin and destination nodes are equal to 1.
%
%\begin{equation}
%\left \{
%\begin{array}{c @{\ =\ } cr}
%    \sum\limits_{i | (i,j) \in \mathcal{E}} p_{ij} & \sum\limits_{k | (j,k) \in \mathcal{E}} p_{jk},& \forall j \in \mathcal{N} \setminus \{n_0,n_d\}\\
%    \sum\limits_{j | (n_0,j) \in \mathcal{E}} p_{n_0j} & \ \ 1&\\
%    \sum\limits_{j | (j,n_d) \in \mathcal{E}} p_{jn_d} & \ \ 1&\\
%\end{array}
%\right.
%\label{eq:flow_constraints}
%\end{equation}
%
This problem is solved as a linear optimization problem by introducing a variable $z$ constrained as follows.
\begin{equation}
\begin{array}{c @{\ \geq\ } lr}
z\ & \sum\limits_{i | (i,j)\in \mathcal{E}} p_{ij} \alpha_j & \forall j \in \mathcal{N}
\end{array}
\label{eq:z_geq_sum}
\end{equation}  

Rewriting Equations (\ref{eq:p_opt_formulation}) and (\ref{eq:z_geq_sum}), the problem can be posed as a linear problem, with $A$ and $b$ representing the flow conservation constraints.
\begin{equation}
\begin{array}{|rl|}
\hline
\multicolumn{2}{|c|}{$minimize$\ z}\\
$subject to$\ \ \ \ \D\p - $\textbf{1}$z & \leq 0\\
\A\p & = \bb \\
\p & \geq\ $\textbf{0}$ \\ \hline
\end{array}
\label{eq:complete_math_formulation}
\end{equation}

\section{\label{sec:framework}Framework for unstructured environments}

%\nblue{maniere dont on modelise + questions que ca pose + sensibilite + delaunay, justifier why}
Though the approach described in Section \ref{sec:approach} has been partially tackled in \cite{Farmey-thesis,joseph-feron}, the previous studies assumed prior existence of a network to optimize the routing on. Using this type of model on different environments requires investigating what a good representation of the environment is. This section presents three methods to build the network that supports the optimization. The outcome of the strategic optimization is compared for different representation of the environment and some key features are identified.

%As it will be explained in section \ref{sec:risk_section}, there are a lot of different factors that can be used to assess the possible casualties at a given location. The problem as described in section \ref{sec:approach} can been applied on road networks for unrealistic applications, as in \cite{Farmey-thesis}. In this part, the question of applying this framework to unstructured environments is tackled. Different network construction methods are described.

	%%%%%%%%%%%%%%%%%%%%%%%%%%%%%%%%%%%%%%%%
	\subsection{\label{subsec:framework_network}Network Construction}

%Assuming the existence of a network is a good hypothesis in most path planning situations. However it limits the vehicle possible routes and therefore decreases the advantage of this approach that is to increase variability to the convoy's trajectory. Moreover, most ambush situations nowadays take place in military situations for which the vehicle concerned are more likely to be TOUT-TERRAIN or aerial vehicle. Therefore it is plausible that adapting this approach as to create a graph representation of the environment, adapted to the vehicle physical model and environment characteristics would be more efficient.	

%Though the approach described in \ref{sec:approach} has been partially tackled in \cite{Farmey-thesis}, the previous studies assumed prior existence of a network to optimize the routing on and is not easily adaptable to new environments. An attempt to address these issues is developed in this section. 

Assuming the existence of a network limits the number of possible routes for the vehicle. Therefore, it decreases the advantage of this approach which aims at increasing the variability of the convoy's trajectory, in particular for vehicles with off-road capabilities. Hence it is expected that expanding the structured network with a graph representation of the unstructured environment will improve the approach efficiency. This representation has to be adapted to the vehicle physical model and environment characteristics.	
	
Several methods are tested in order to create a high fidelity directed network with respect to the environment while allowing reasonably short computation time. Differences were made on the sampling method and on the connectivity between nodes, as displayed in Table \ref{method_table}. While randomly sampled nodes connected through a Delaunay triangulation result in a relatively small and computationally efficient representation of the environment, \rdm might not be representative enough of the details of the environment. The most complete representation of the environment is obtained through \unih, but it requires 16 directed edges per node (8 neighboring nodes, two directions per edge). This method is computationally intensive, therefore it might be preferable to choose a less precise technique with yet enough precision on the environment description. These methods will be compared in Section \ref{sec:performance}.
%\red{ERIC: should i explain Delaunay triangulation and a 8-connected grid?}

\begin{table}
\begin{center}
$\begin{array}{|c|c|c|}
\hline
$Method \#$ & $Sampling$ & $Connectivity$ \\
\hline
$1$ (\texttt{rdm}) & $Random$ & $Delaunay triangulation$\\
$2$ (\unih) & $Uniform$ & $8 connected grid$\\
$3$ (\uniD) & $Uniform$ & $Delaunay triangulation$\\
\hline\\
\end{array}$
\caption{Different network construction methods.}
\label{method_table}
\end{center}
\end{table}
	
	%%%%%%%%%%%%%%%%%%%%%%%%%%%%%%%%%%%%%%%%
	\subsection{Ambush types}
A different model is now considered where the state space for RED is modified. Ambushes are paired with an area in the environment instead of a node on the network.
Since the initial purpose of this framework %\red{(back when Farmey \cite{Farmey-thesis} was doing it)} 
was to create routes minimizing losses on existing networks, considering point-based ambushes was a starting point. Now, the case of fixed geographic ambush areas is considered. This model is more realistic in the sense that it relays the idea of a maximum reach for RED, which defines the size of the ambush areas.
Another important reason to have ambush by area is that it allows to decouple the space dependency for BLUE strategy from RED's strategy in order to study the convergence of BLUE's strategy regarding the precision of the environment model when the set of ambushes for RED is fixed. This is a very interesting feature if one wishes to use the discrete routing strategies as approximations of a continuous strategy. The results will give us insights on the possibility to construct such a continuous solution.

For this case, a transposition matrix $\textbf{S}$ is created. $\textbf{S}$ describes the belonging of a node to an ambush area: if node $j$ belongs to ambush area $i$ then the $j^{th}$ column of $\textbf{S}$ will be zero except on its $i^{th}$ line. This notation allows the reformulation of the linear problem to be very close to the initial formulation. BLUE's optimal solution is now: $\p^{\ast} = \arg\min\limits_{\p} \max\limits_{\q} \q^t \textbf{S}\D \p$.

%\red{ERIC: This S is the one currently implemented. It is however false because we should only be interested in edges entering an ambush area and not all edges related to a node inside an ambush area (nodes in the middle of an ambush area are useless). It is interesting because starting at some resolution, the strategy for BLUE could be summarized on a reduced network of 8-connected ambush areas.}

\begin{figure}[!ht]
\begin{center}
\subfigure[Area size = 6]{\label{fig:simple_area_with_q1}\includegraphics[width=0.20\textwidth, height=0.20\textwidth]{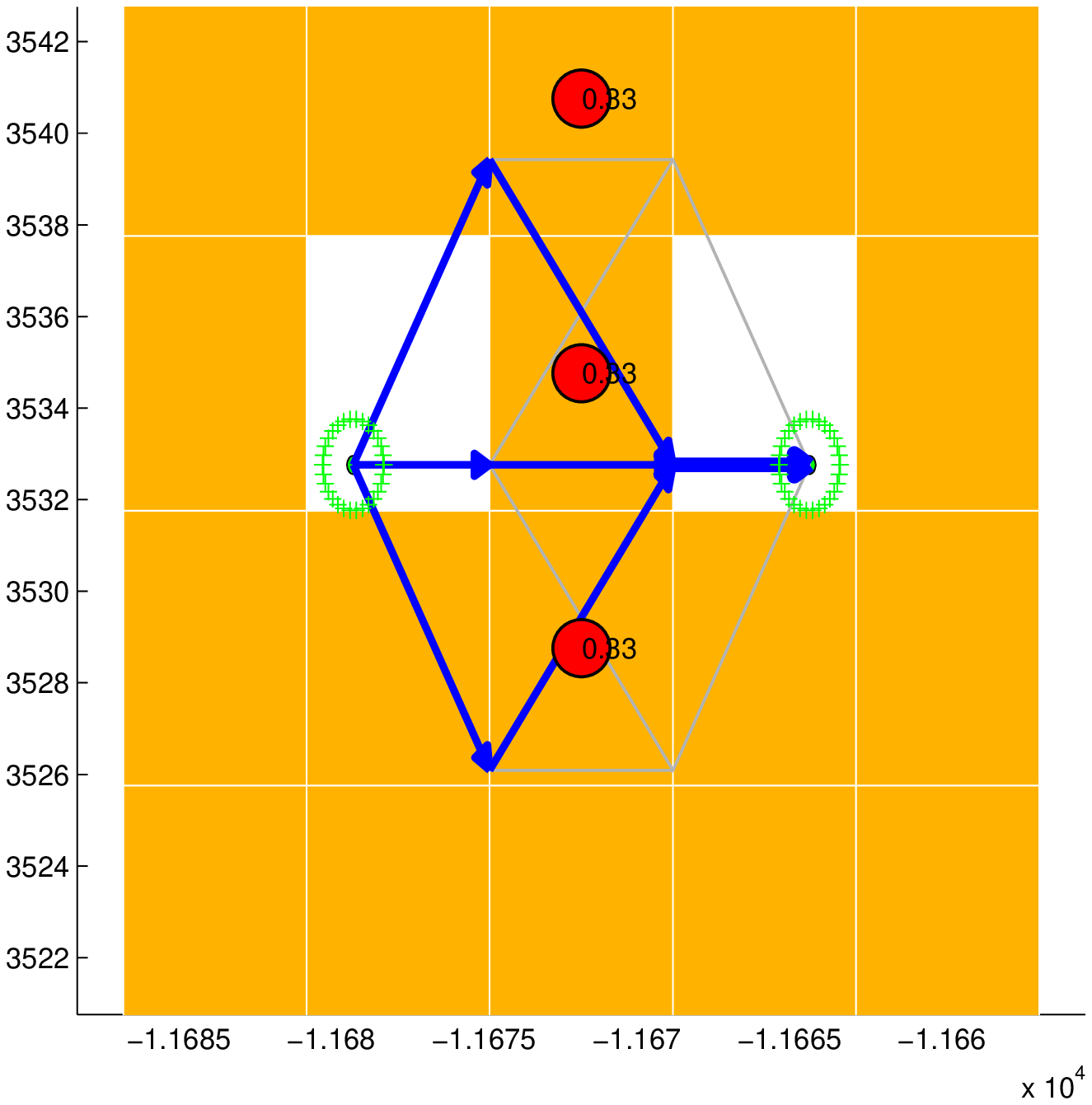}}\qquad
\subfigure[Area size = 10]{\label{fig:simple_area_with_q2}\includegraphics[width=0.20\textwidth, height=0.20\textwidth]{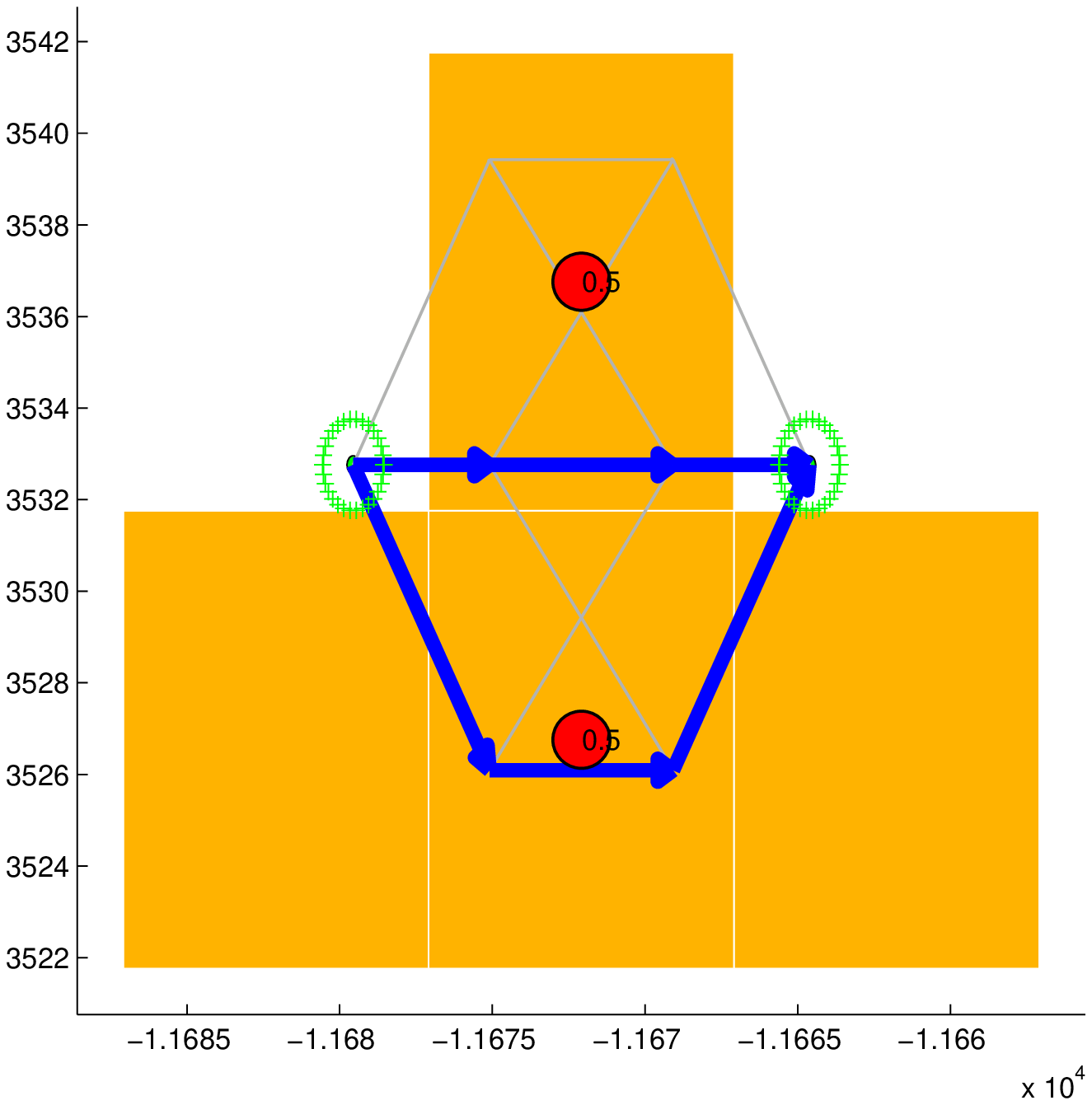}}%
\caption{Simple network example - Ambushes by area. The probability associated width each edge is displayed as the width of the edge. The orange rectangles represent the intensity of the local outcome on corresponding ambush areas. The origin and destination area are supposed to have a zero local outcome. Grey edges are unused or used with very low probability. Red circles represent the strategy chosen by RED, ie areas $i$ where RED might set his ambush with probability $q_i>0$. This example illustrates the importance of associating an area to an ambush. Increasing the reach of RED between (a) and (b) leads to a drastically different strategy: the reach of RED has very important consequences on BLUE's strategy.}
\label{fig:simple_area_with_q}
\end{center}
\end{figure}

Looking back at the trivial example, two examples of ambush area sets are displayed in Figure \ref{fig:simple_area_with_q}. As reflected in this Figure, the reach of RED has a great influence on BLUE's optimal strategy. This validates our motivation to include this parameter in the environment model. Note that the strategy $\q^*$ for RED is computed assuming it has perfect knowledge of BLUE's strategy. This means that RED knows the network and the probability that BLUE uses any edge of the network. Once $\p^*$ is computed for BLUE, $\q^*$ is computed as $\q^* = \arg \max \limits_{\q} \q^t \textbf{S}\D \p^*$.
This is a very strong assumed advantage for RED regarding which this routing method is quite robust.

%\begin{figure}[!ht]
%\begin{center}
%\subfigure[Area size = 6]{\label{fig:simple_area1}\includegraphics[width=2in,height=2in]{simple_network_sol_area1.eps}}\qquad
%\subfigure[Area size = 10]{\label{fig:simple_area2}\includegraphics[width=2in,height=2in]{simple_network_sol_area2.eps}}%
%\caption{Simple network example - Ambushes by area. The ambush areas are delimited by the red dashed lines. This example illustrates the importance of associating an area to an ambush. The reach of RED has very important consequences on BLUE's strategy.}
%\label{fig:simple_area}
%\end{center}
%\end{figure}

	%%%%%%%%%%%%%%%%%%%%%%%%%%%%%%%%%%%%%%%%
	\subsection{Energy Optimization}

%\nblue{avoid cycles in graph so that our route does not loop indefinitely}

A consequence of taking into account the reach of RED is the creation of cycles inside each ambush area. This means that the probabilistic routing could lead to a path where the vehicle stays inside a closed subspace for a very long time. This situation is more likely in ambush areas where the local outcome is close to zero.
This illustrates one of the issues with this routing method, which is the absence of any kind of energy optimization. 

In order to avoid cycles, an energy optimization factor $\lambda$ is introduced that allows to tune the outcome of the optimization. Large values of $\lambda$ will lead to a few different paths close to the shortest path while increasing the overall risk for the system. Low values of $\lambda$ will return routing strategies close to the safest ones. Note that, the set of optimal strategy being quite large, the routing strategies returned with low values of $\lambda$ might also be risk optimal in several situations. The new objective function is $(1-\lambda)z + \lambda \sum\limits_{i,j | (i,j) \in \mathcal{E}} p_{ij}\|e_{ij}\|$.

%\nblue{see math optimization change}

%
%\begin{equation}
%\begin{array}{rl}
%%\multicolumn{2}{c}{$minimize$\ (1-\lambda)z + \lambda \sum\limits_{i,j | (i,j) \in \mathcal{E}} p_{ij}\|e_{ij}\|}\\
%$subject to$\ \ \ \ \D\p - $\textbf{1}$z & \leq 0\\
%\A\p & = \bb \\
%\p & \geq\ $\textbf{0}$
%\end{array}
%\label{eq:math_with_energy}
%\end{equation}
	
\begin{figure}[!ht]
\begin{center}
\subfigure[\texttt{rdm} - $\lambda=0$.]{\label{fig:rdm_100_noE}\includegraphics[width=0.13\textwidth, height=0.13\textwidth]{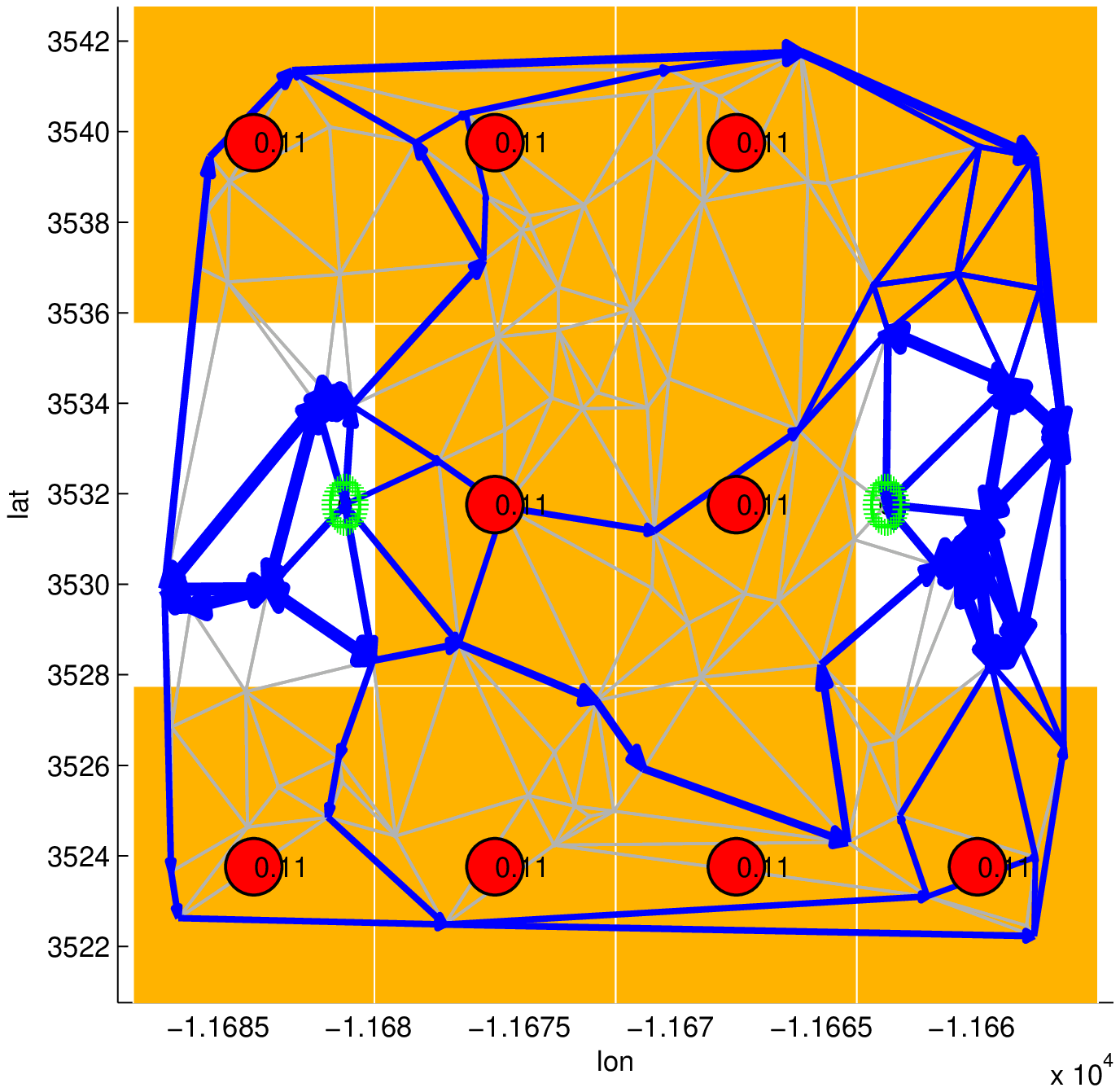}}\qquad
\subfigure[\texttt{uni8} - $\lambda=0$.]{\label{fig:uni8_100_noE}\includegraphics[width=0.13\textwidth, height=0.13\textwidth]{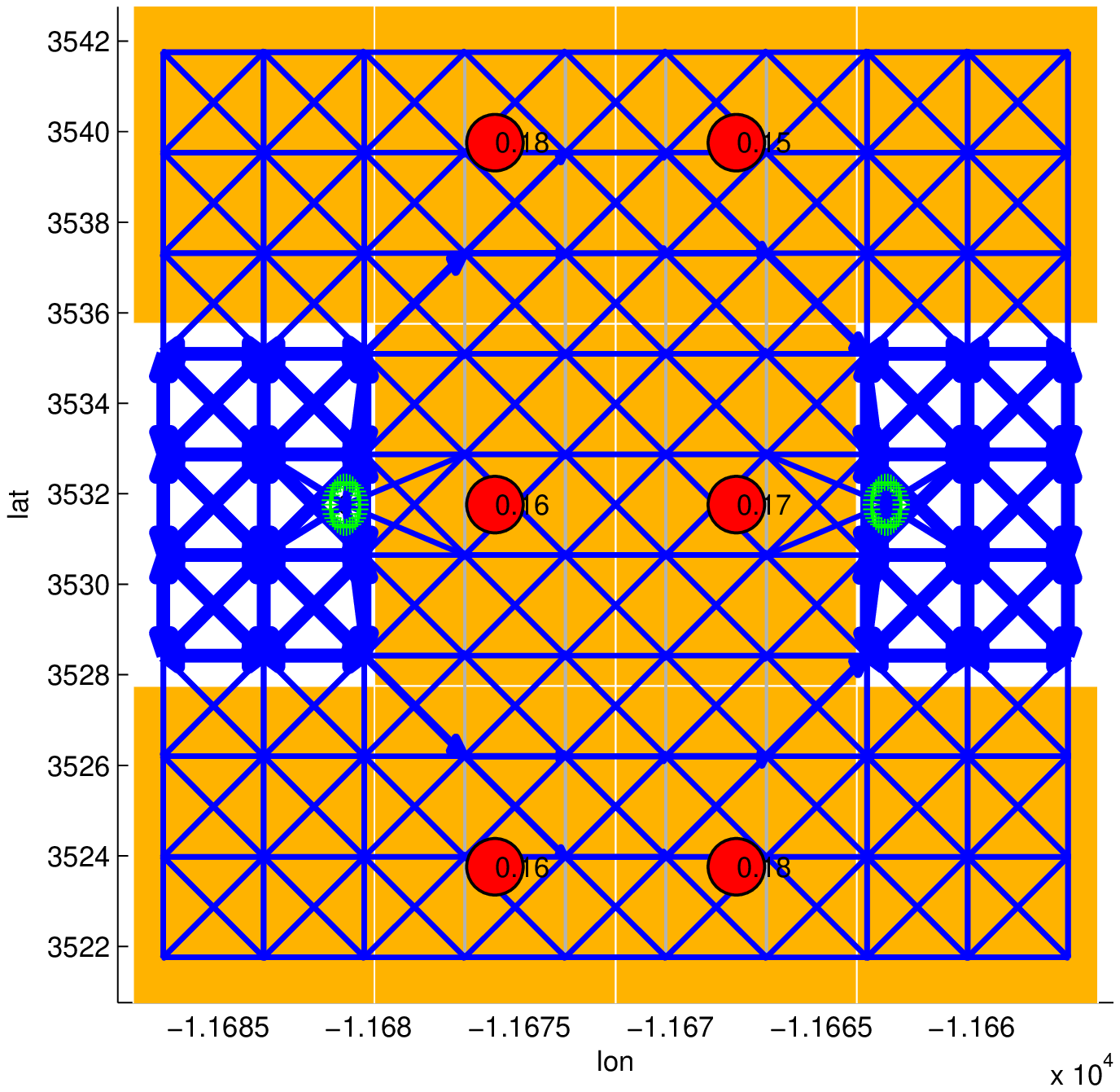}}\qquad
\subfigure[\texttt{uniD} - $\lambda=0$.]{\label{fig:uniD_100_noE}\includegraphics[width=0.13\textwidth, height=0.13\textwidth]{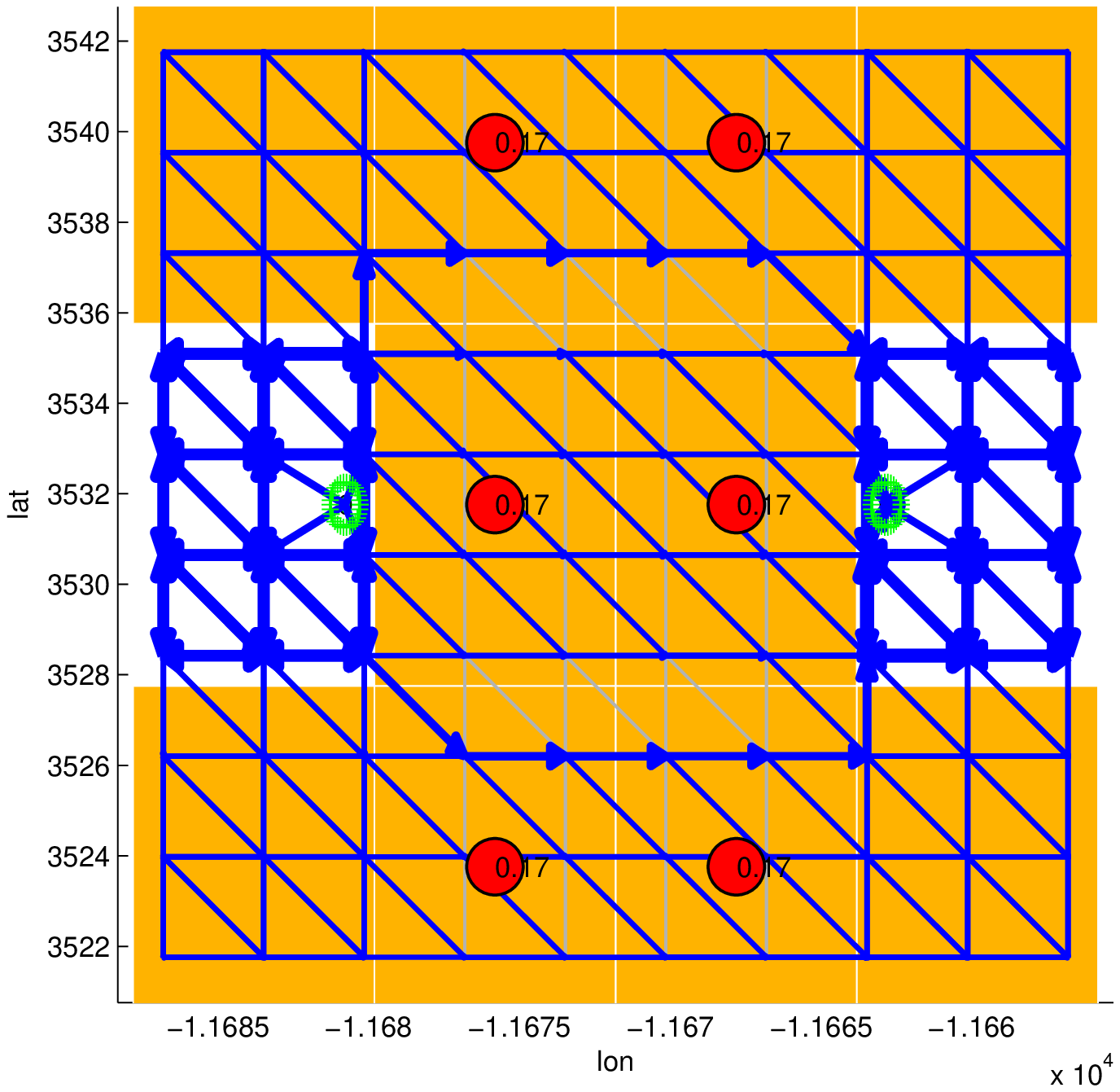}}\\
\subfigure[\texttt{rdm} - $\lambda=10^{-4}$.]{\label{fig:rdm_100}\includegraphics[width=0.13\textwidth, height=0.13\textwidth]{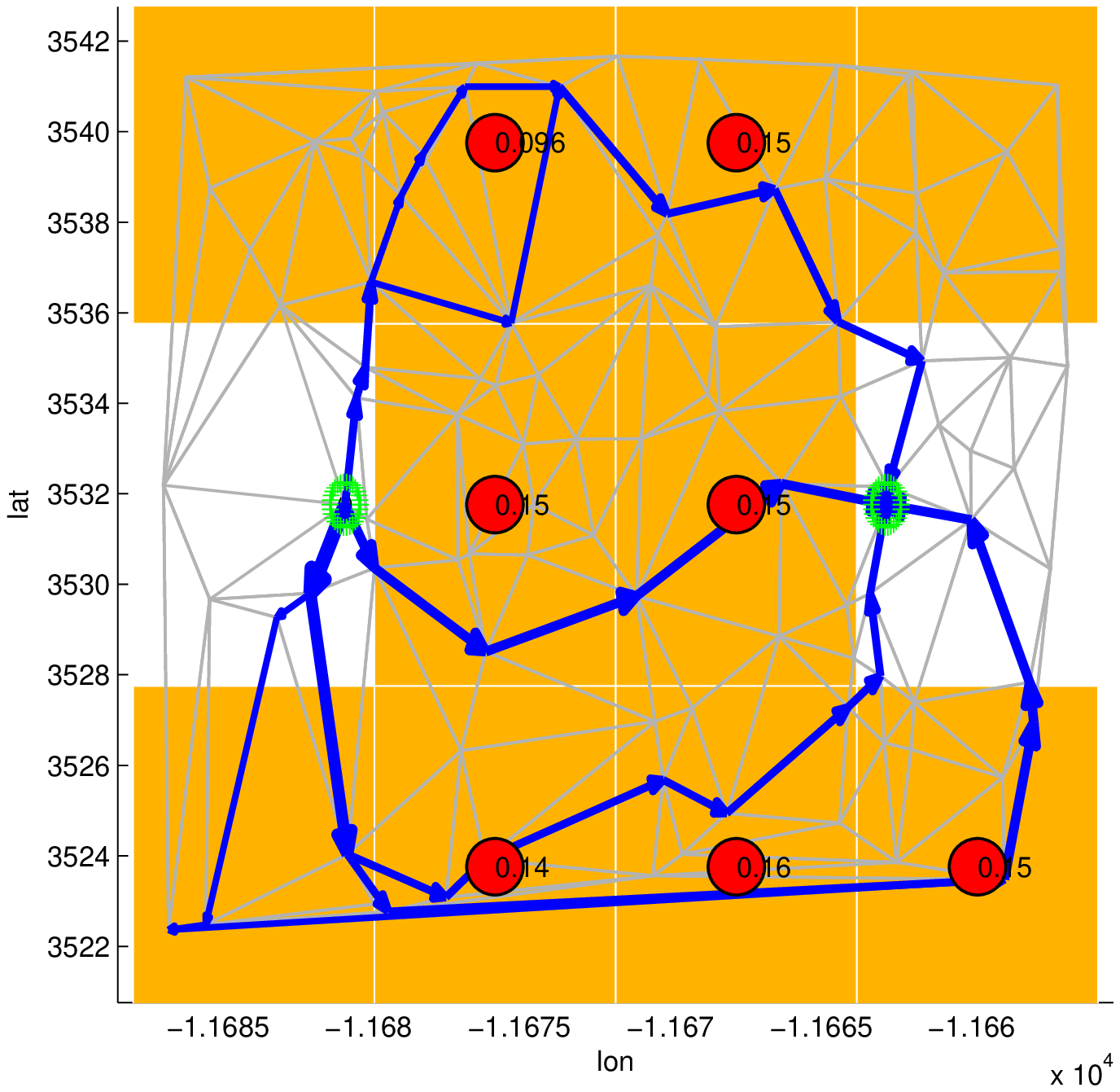}}\qquad
\subfigure[\texttt{uni8} - $\lambda=10^{-4}$.]{\label{fig:uni8_100}\includegraphics[width=0.13\textwidth, height=0.13\textwidth]{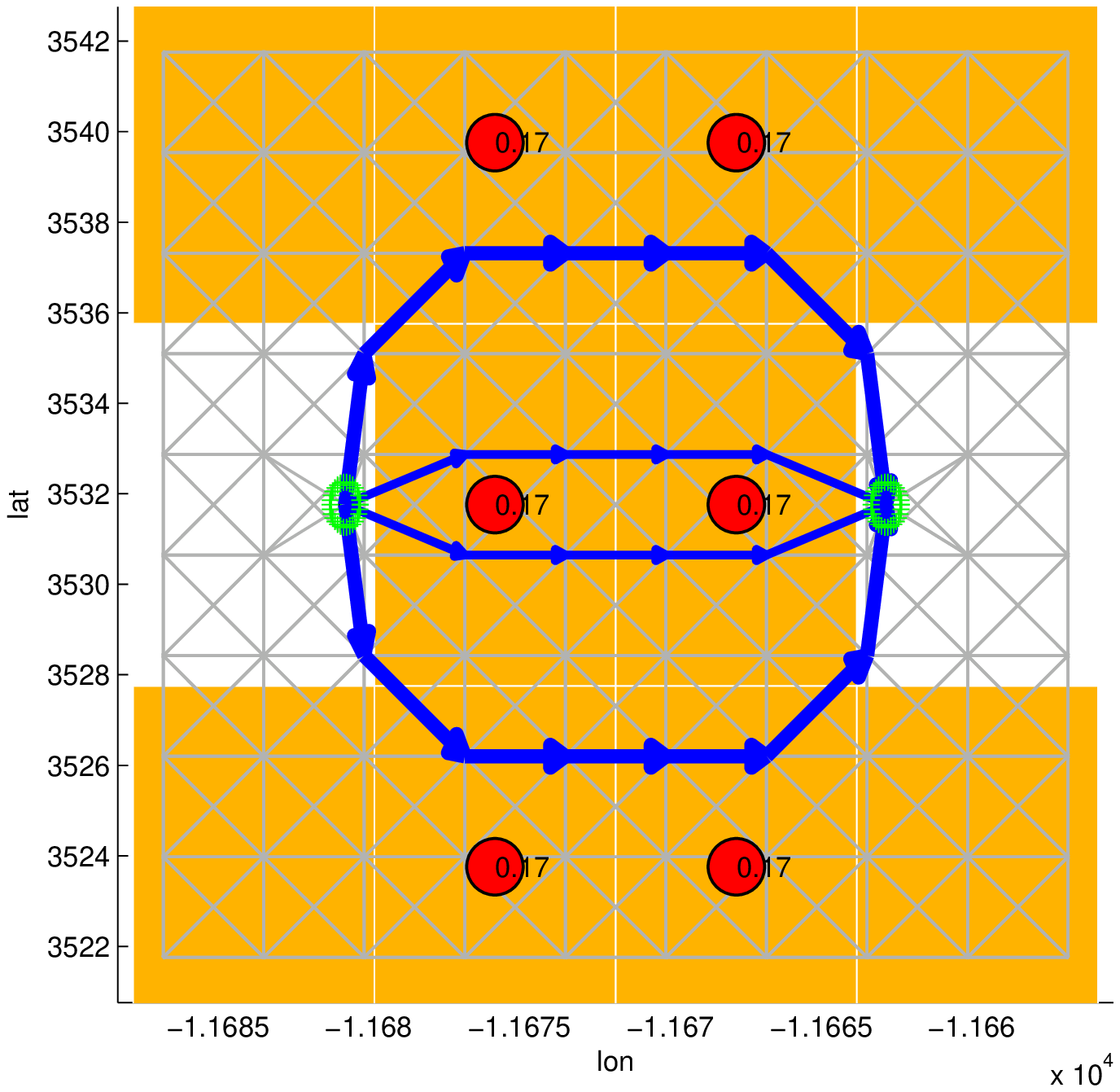}}\qquad
\subfigure[\texttt{uniD} - $\lambda=10^{-4}$.]{\label{fig:uniD_100}\includegraphics[width=0.13\textwidth, height=0.13\textwidth]{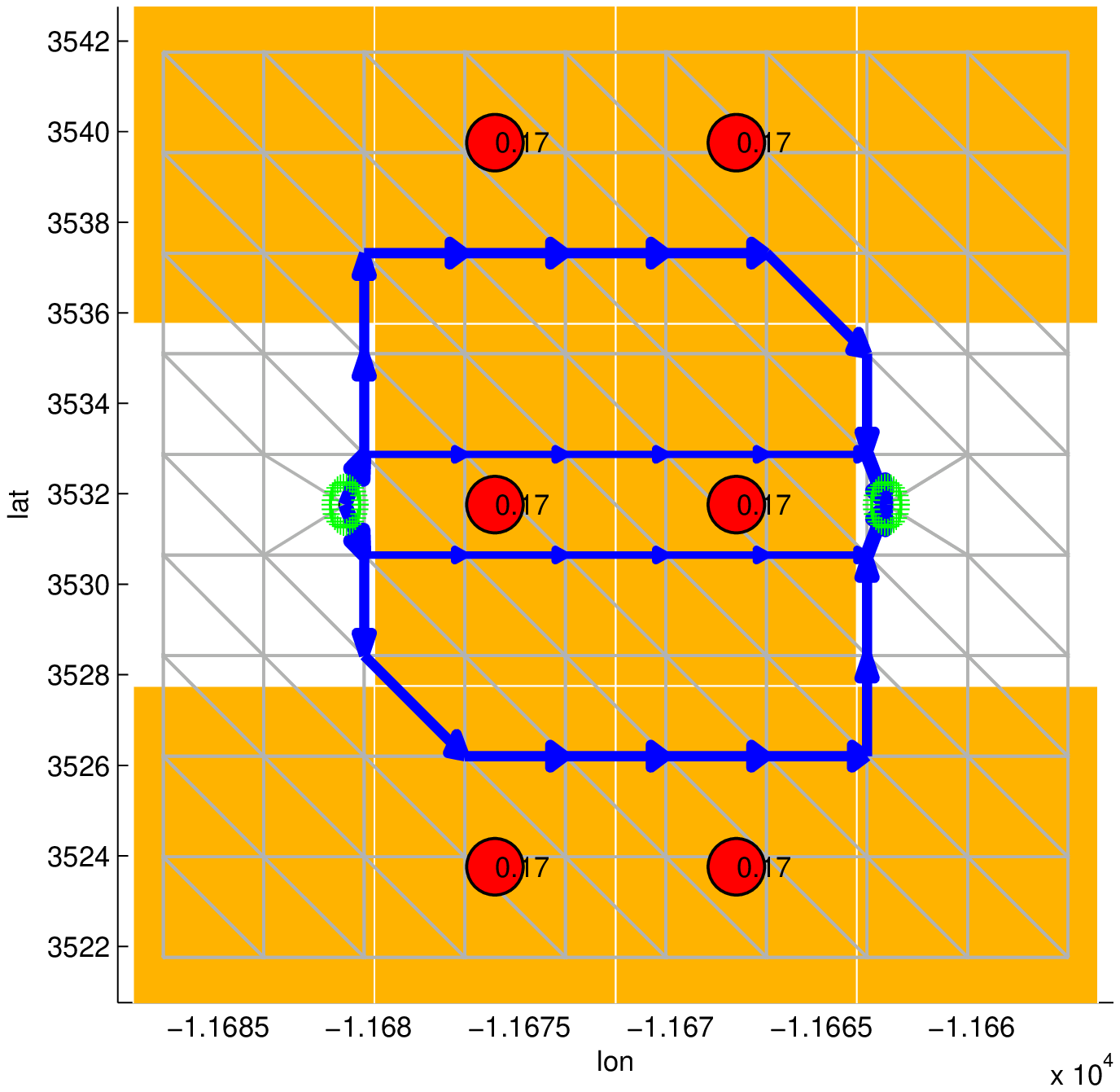}}%%
\caption{Results of different network construction methods. In the left column, cycles can be identified in the departure and arrival areas whereas there are only outgoing edges in the right column examples. An interesting property of the route proposed with energy optimization is the 1-to-1 repartition of route segments to areas (as seen in Figures \ref{fig:uni8_100} and \ref{fig:uniD_100}) along the median between departure and arrival.}
\label{fig:3methods_100}
\end{center}
\end{figure}

%\nblue{1-to-1 mapping gives 7 paths}
%\nblue{on uni8 and uniD red optimal strategy is to spread across the entire environment}
The optimal routes returned by our framework for different network construction method can be first analyzed through their graphical representation.
Examples of routing strategies with low or no energy optimization for different network construction methods described in \ref{subsec:framework_network} are compared in Figure \ref{fig:3methods_100}. 
It is important to note that the routes with and without energy optimization are very much alike. Therefore the cycles are suppressed without significant change to the routing strategy regarding the outcome of the game. The energy optimization also allows to avoid a side effect where flows agglomerate on the border of the environment.

Interestingly, it seems that the route includes a fixed number of distinct paths depending on the reach of RED. In Figure \ref{fig:3methods_100}, this distance is such that the median between departure and arrival is divided into three segments. This leads to four distinct paths in Figures \ref{fig:uniD_100},\ref{fig:uni8_100}. Yet, this is optimally equivalent to having only one path along the set of areas aligned between departure and arrival. In fact any route where paths do not intersect with a uniform flow distribution along the ambush areas would be optimally equivalent regarding the strategic outcome.
Finally the network construction method seem to have an important impact on RED strategy. Comparing Figures \ref{fig:rdm_100_noE},\ref{fig:rdm_100} and Figures \ref{fig:uni8_100_noE},\ref{fig:uni8_100},\ref{fig:uniD_100_noE},\ref{fig:uniD_100}, RED has a strategy with a larger spread on grid-based networks. This could lead us to conclude that these networks allow for better results. This is however not the case because the strategy displayed in Figures \ref{fig:uni8_100_noE},\ref{fig:uni8_100},\ref{fig:uniD_100_noE},\ref{fig:uniD_100} is similar to a single ambush area strategy with probability $1.0$ across any area on the departure-arrival median.

%\nblue{Meaning of above?}

%%%%%%%%%%%%%%%%%%%%%%%%%%%%%%%%%%%%%%%%%%%%%%%%%%%%%%
\section{\label{sec:performance}Performance}

	%%%%%%%%%%%%%%%%%%%%%%%%%%%%%%%%%%%%%%%%
	\subsection{Metrics description}

%\begin{figure}[!ht]
%\begin{center}
%\includegraphics[width=3in,height=1.8in]{entropy_1D.png}
%\label{fig:entropy_1d}
%\caption{Different solution distribution and corresponding entropy for the 1D ambush example.}
%\end{center}
%\end{figure}

In order to assess the performances of the optimization, four metrics are used. In his study of the one dimensional problem, Ruckle identified the uniform distribution of paths along a line as the optimal strategy. This strategy corresponds to maximizing the entropy of the probability distribution. Based on this idea, the probability of crossing along the median between departure and arrival is computed. 
The \textit{entropy} is used as the first metric, it is high for deceptive routes and close to zero for non-deceptive ones. 
The idea of the optimal strategy being uniformly distributed on the environment led to the creation of another metric, the \textit{spreading} of the route, which measures the portion of the environment covered by the route as a fraction of the total surface of the environment. For any ambush area the probability of entering this area can be computed as the sum of inward flows. If the area is reached with probability $p> p_{min}$, the surface of this area is computed in the spreading. 
The \textit{strategic outcome} $V$ as computed in (\ref{eq:strategic_outcome}) is one of the most important metrics. Recall that $V$ accounts for the expected losses for BLUE during one trip. It reflects how well the route can be optimized to reduce these losses. 
The \textit{energy} metric is defined as $E= \sum\limits_{i,j | (i,j) \in \mathcal{E}} p_{ij}\|e_{ij}\|$. This metric represents the divergence of the stochastic strategy from the shortest path between origin and destination.

	%%%%%%%%%%%%%%%%%%%%%%%%%%%%%%%%%%%%%%%%
	\subsection{Results}
	
%\red{Network construction method sensitivity}		
	
\begin{figure}[!ht]
\begin{center}
%\subfigure[Network description.]{\label{fig:rdm_900}\includegraphics[width=1.6in, height=1.4in]{rdm_900nodes_alpha10000_energy0_0001_obs0_intpoint_areaAmbush2_79_path.png}}\qquad
%\subfigure[Network description.]{\label{fig:rdm_3600}\includegraphics[width=1.6in, height=1.4in]{rdm_3600nodes_alpha10000_energy0_0001_obs0_intpoint_areaAmbush2_79_path.png}}\\
%
\subfigure[900 vertices]{\label{fig:uni8_900}\includegraphics[width=0.20\textwidth, height=0.20\textwidth]{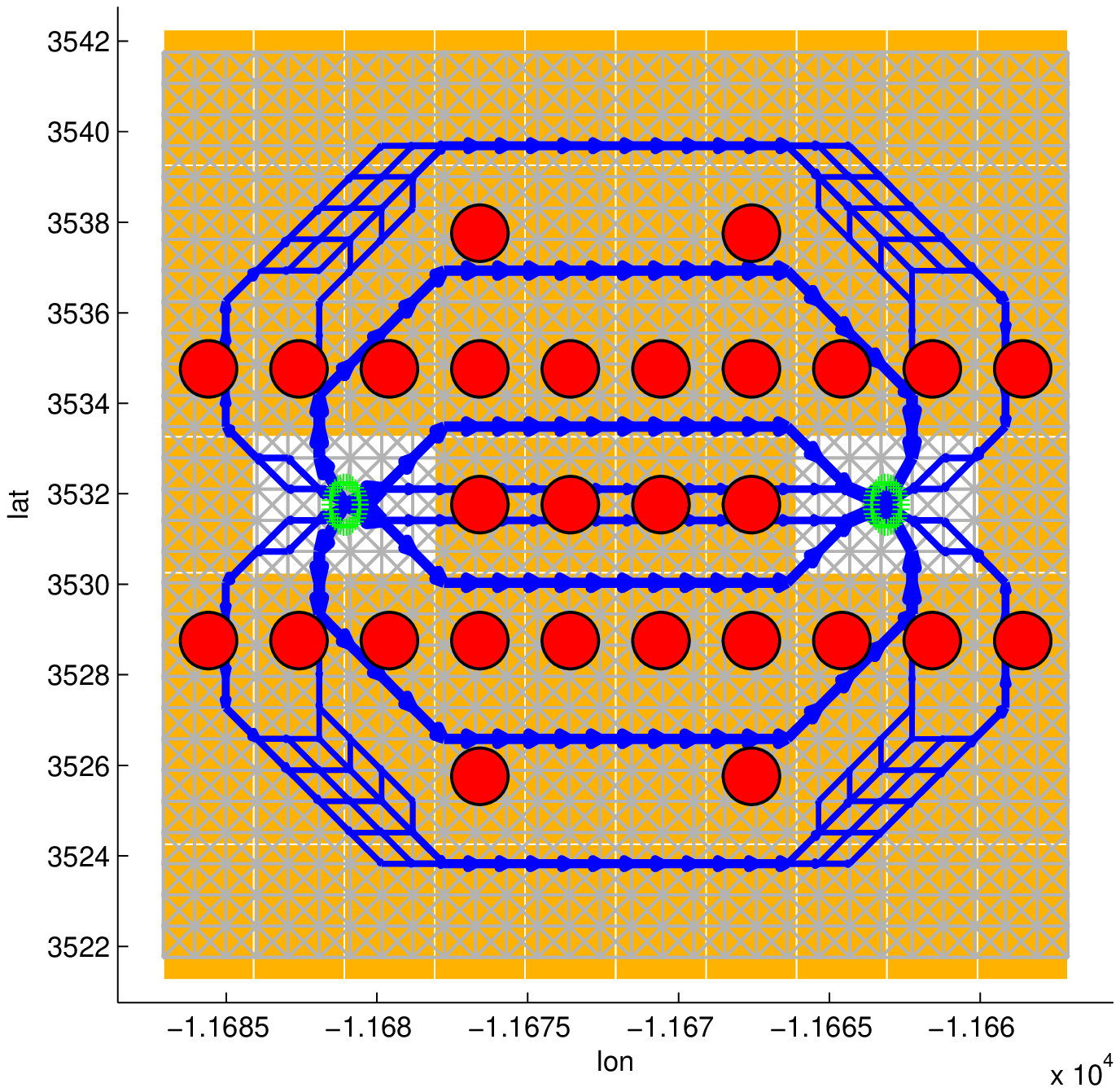}}\qquad
\subfigure[3600 vertices]{\label{fig:uni8_3600}\includegraphics[width=0.20\textwidth, height=0.20\textwidth]{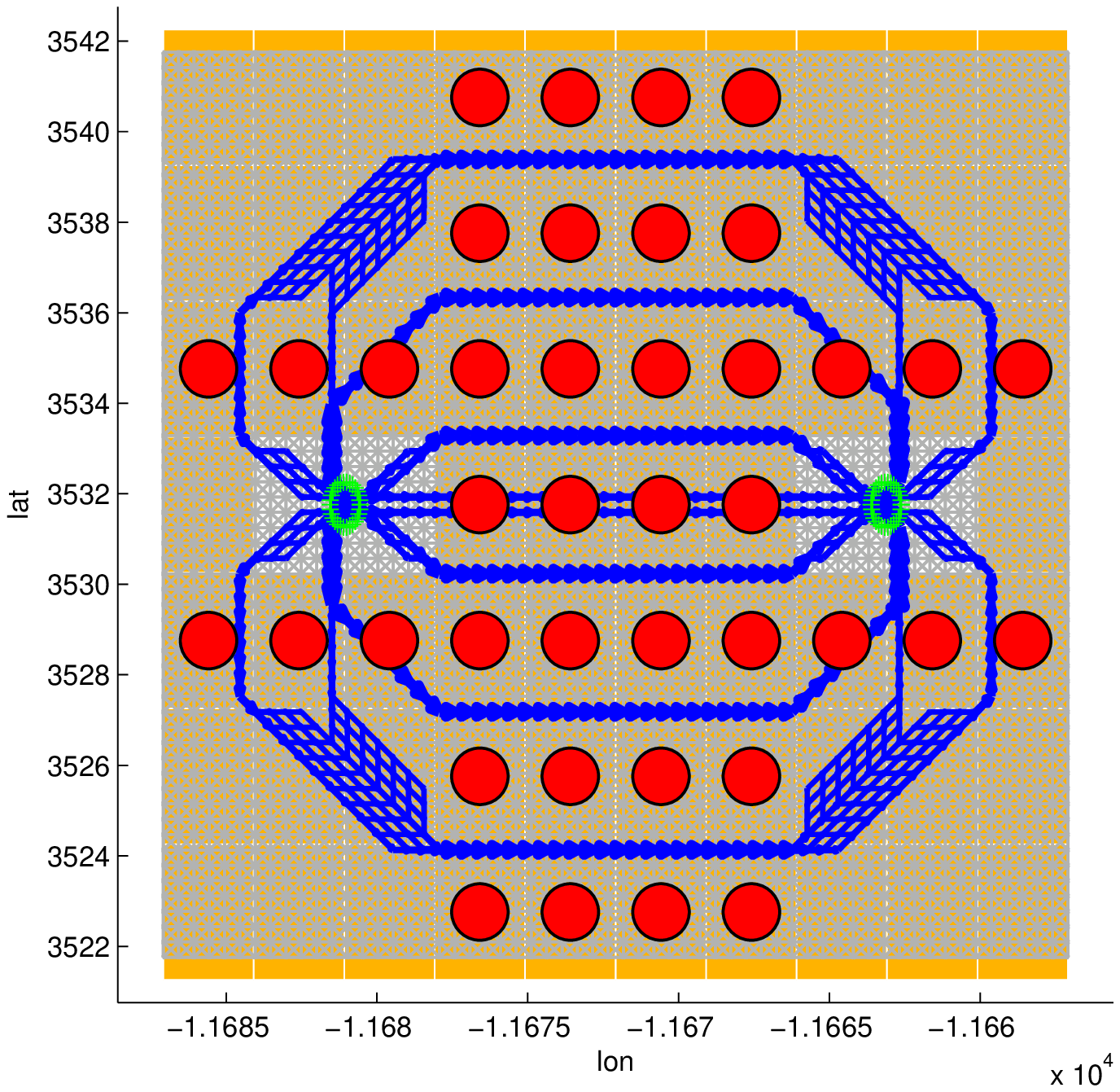}}\\
%
%\subfigure[Network description.]{\label{fig:uniD_900}\includegraphics[width=1.6in, height=1.4in]{uniD_900nodes_alpha10000_energy0_0001_obs0_intpoint_areaAmbush2_79_path.png}}\qquad
%\subfigure[Network description.]{\label{fig:uniD_3600}\includegraphics[width=1.6in, height=1.4in]{uniD_3600nodes_alpha10000_energy0_0001_obs0_intpoint_areaAmbush2_79_path.png}}\\
%
\caption{Comparison of the results obtained for different network density with the second network construction method. The departure-arrival median is divided in seven areas, which leads to seven different paths. The routes in (a) and (b) are sensibly alike, illustrating the convergence of our solution when the size of the network increases.}
\label{fig:convergence_example}
\end{center}
\end{figure}

Figure \ref{fig:convergence_example} illustrates one of the main results of this paper. For each method, BLUE's strategy is converging to a distribution that depends solely on RED's reach and the local outcome. The reach of RED in the following environment is such that the median between the departure and arrival points is divided into seven sections. This leads to an optimal theoretical outcome of $\frac{1}{7} \simeq 0.1456$ and hence a theoretical optimal entropy of $\sum \limits_{i=1}^7 -\frac{1}{7} \log(\frac{1}{7})= \log(7) \simeq 1.94$.

\begin{figure*}[!ht]
\center
\subfigure[Outcome  - $\lambda = 0$]{\includegraphics[width=.30\textwidth,height=.18\textwidth]{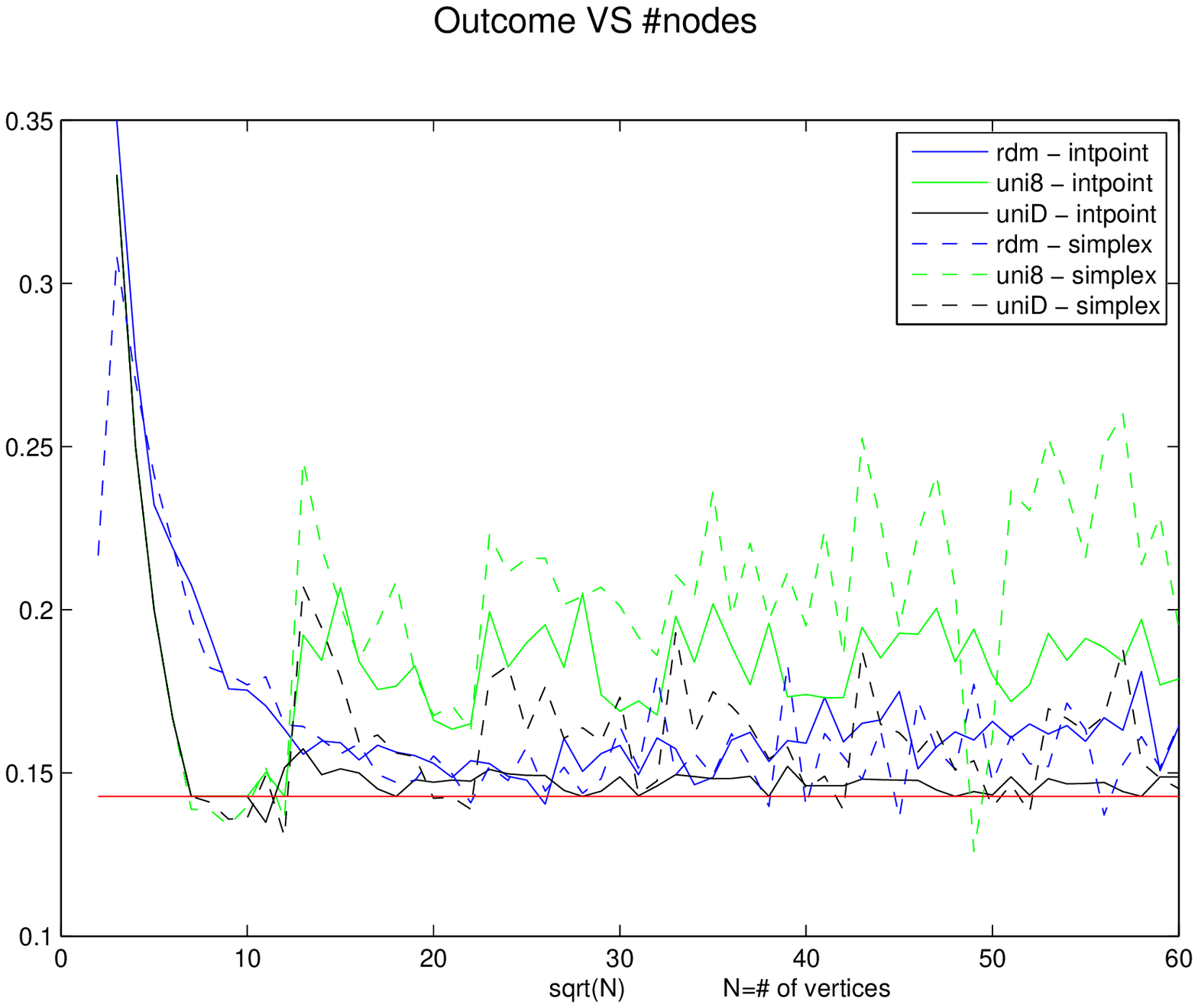}
\label{fig:outcome_vs_nodes_OPTIMIZATION_noE}}\qquad
\subfigure[Outcome  - $\lambda = 10^{-4}$]{\includegraphics[width=.30\textwidth,height=.18\textwidth]{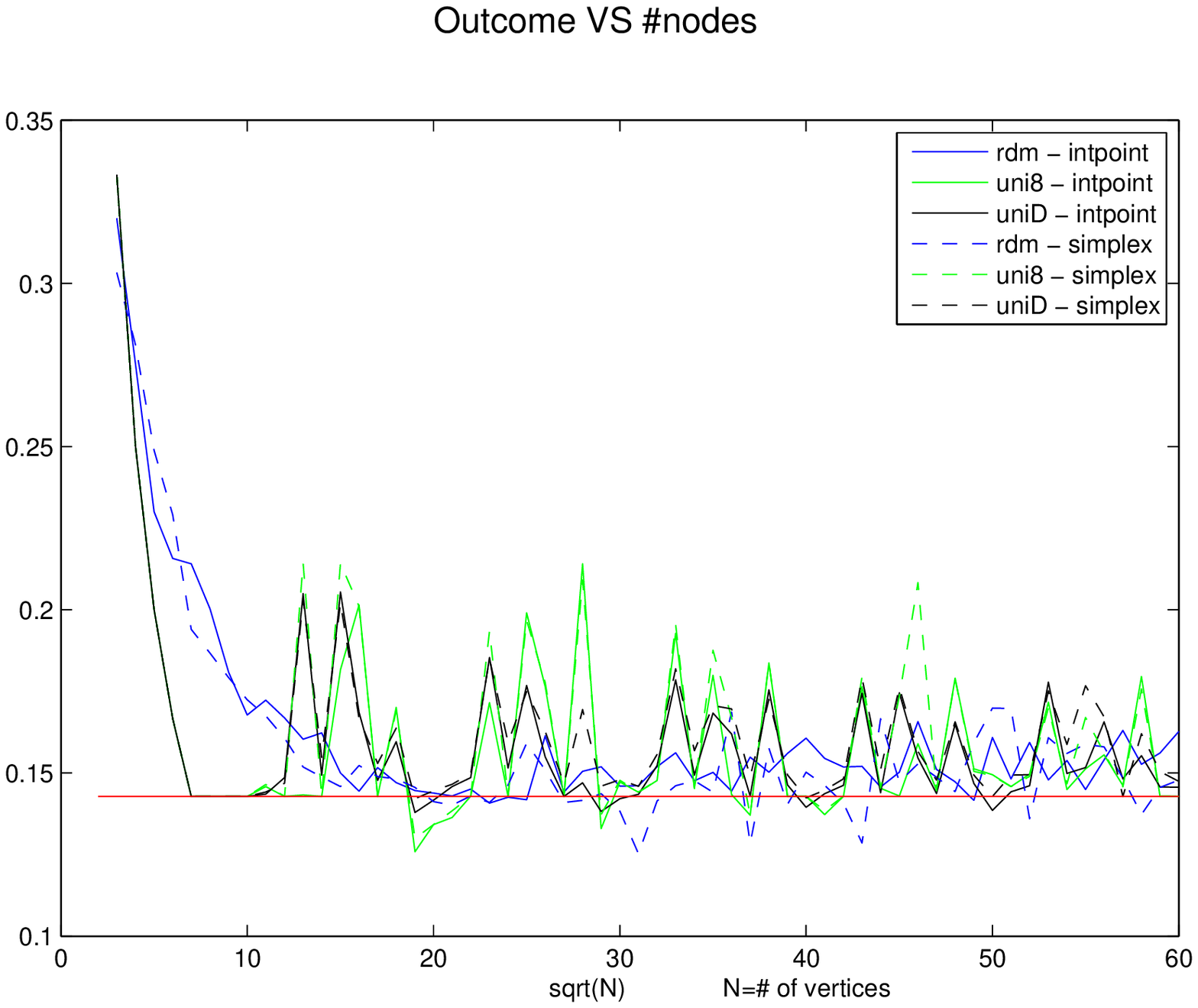}
\label{fig:outcome_vs_nodes_OPTIMIZATION_withE}}\\
\subfigure[Spreading  - $\lambda = 0$]{\includegraphics[width=.30\textwidth,height=.18\textwidth]{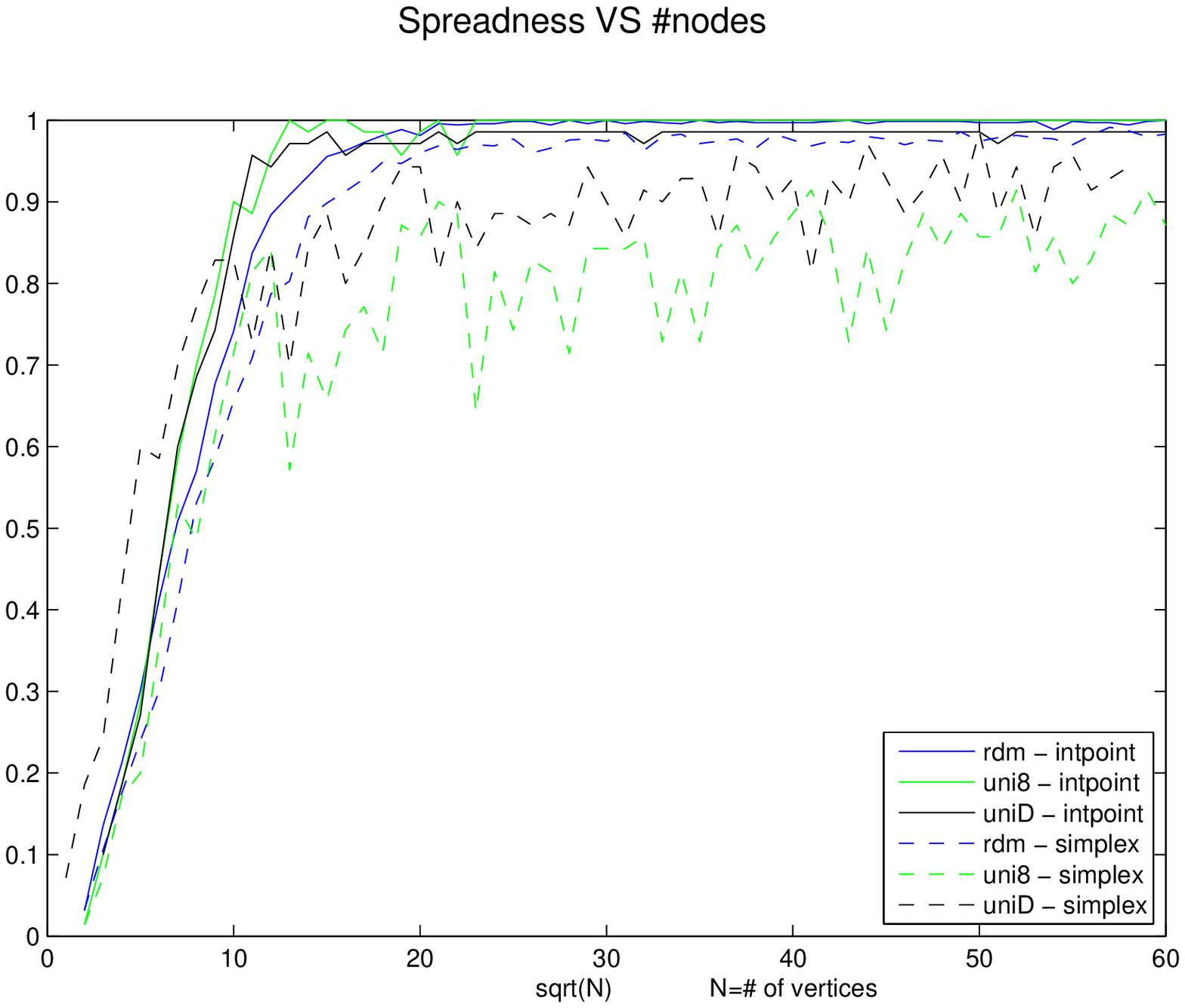}
\label{fig:spreadness_vs_nodes_OPTIMIZATION_noE}}\qquad
\subfigure[Spreading  - $\lambda = 10^{-4}$]{\includegraphics[width=.30\textwidth,height=.18\textwidth]{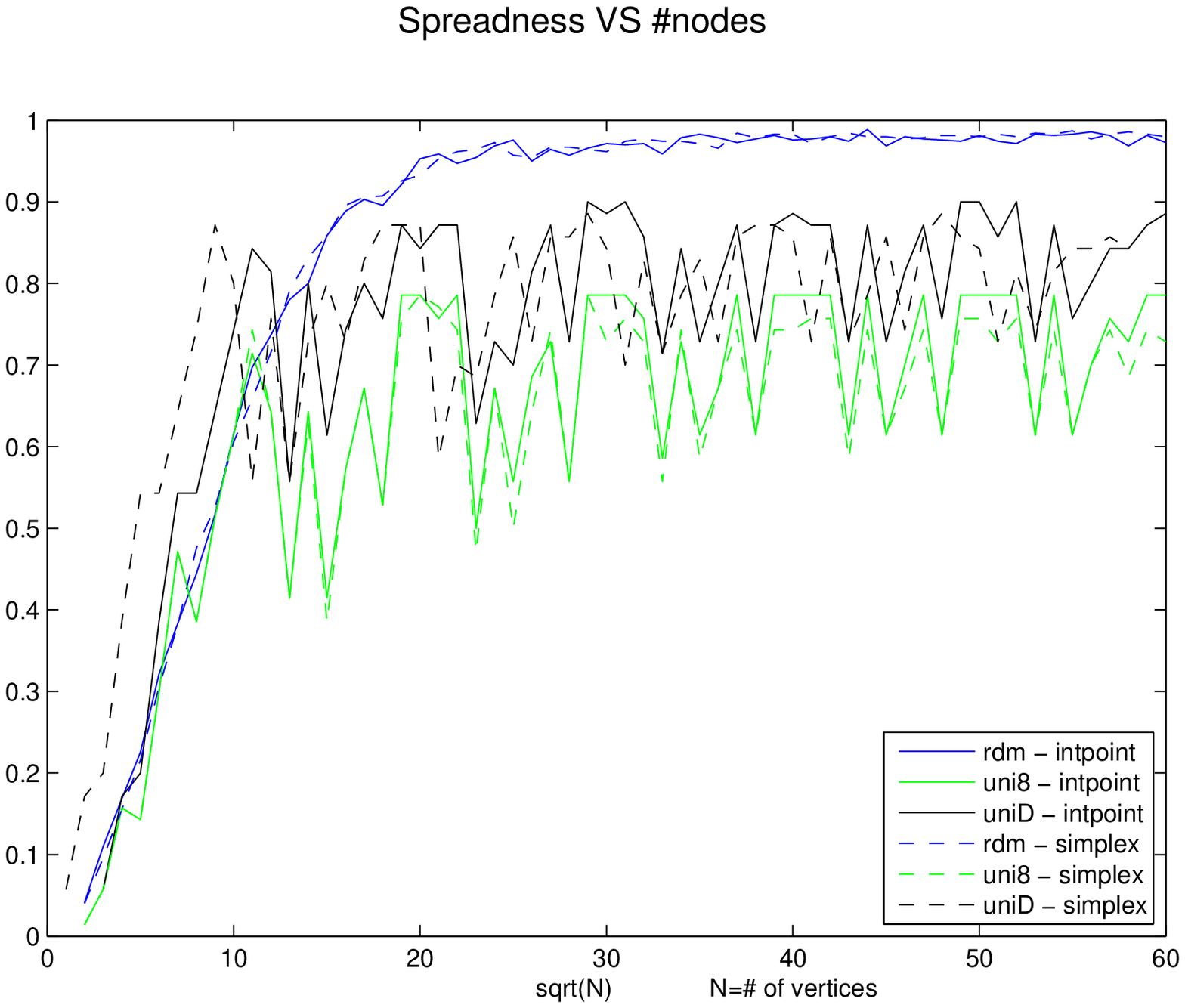}
\label{fig:spreadness_vs_nodes_OPTIMIZATION_withE}}\\
\subfigure[Energy  - $\lambda = 0$]{\includegraphics[width=.30\textwidth,height=.18\textwidth]{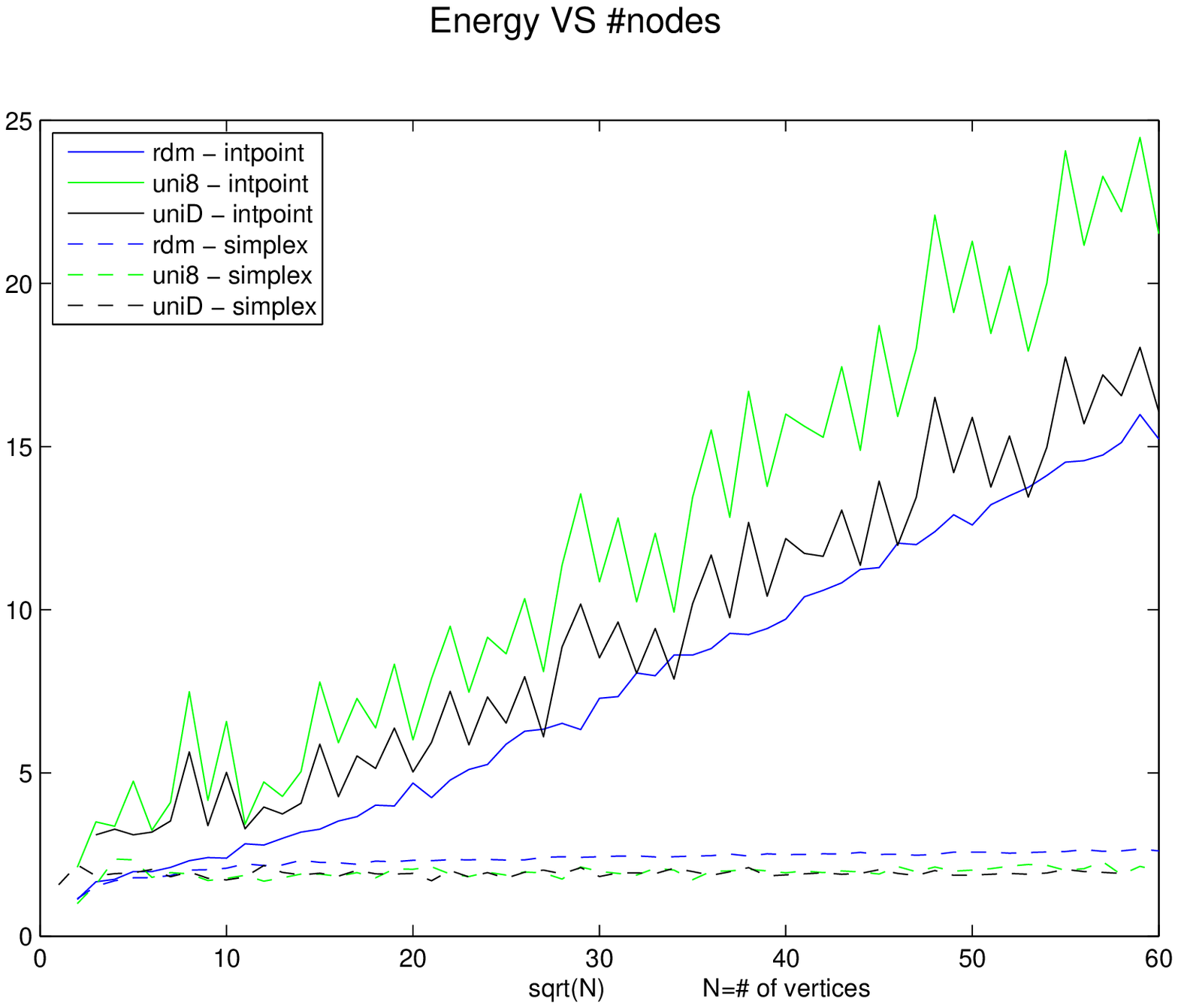}
\label{fig:energy_vs_nodes_OPTIMIZATION_noE}}\qquad
\subfigure[Energy  - $\lambda = 10^{-4}$]{\includegraphics[width=.30\textwidth,height=.18\textwidth]{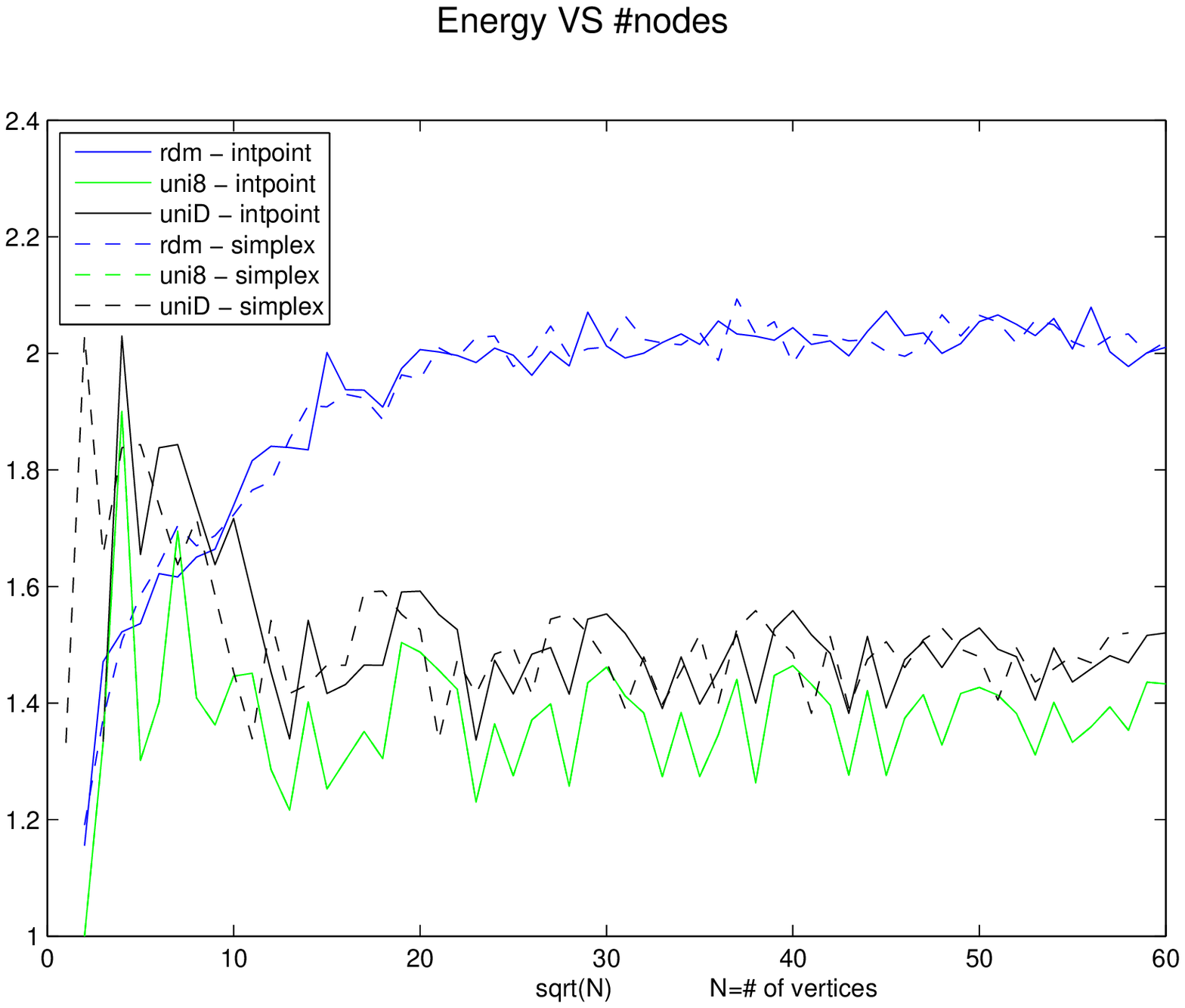}
\label{fig:energy_vs_nodes_OPTIMIZATION_withE}}\\
\subfigure[Entropy  - $\lambda = 0$]{\includegraphics[width=.30\textwidth,height=.18\textwidth]{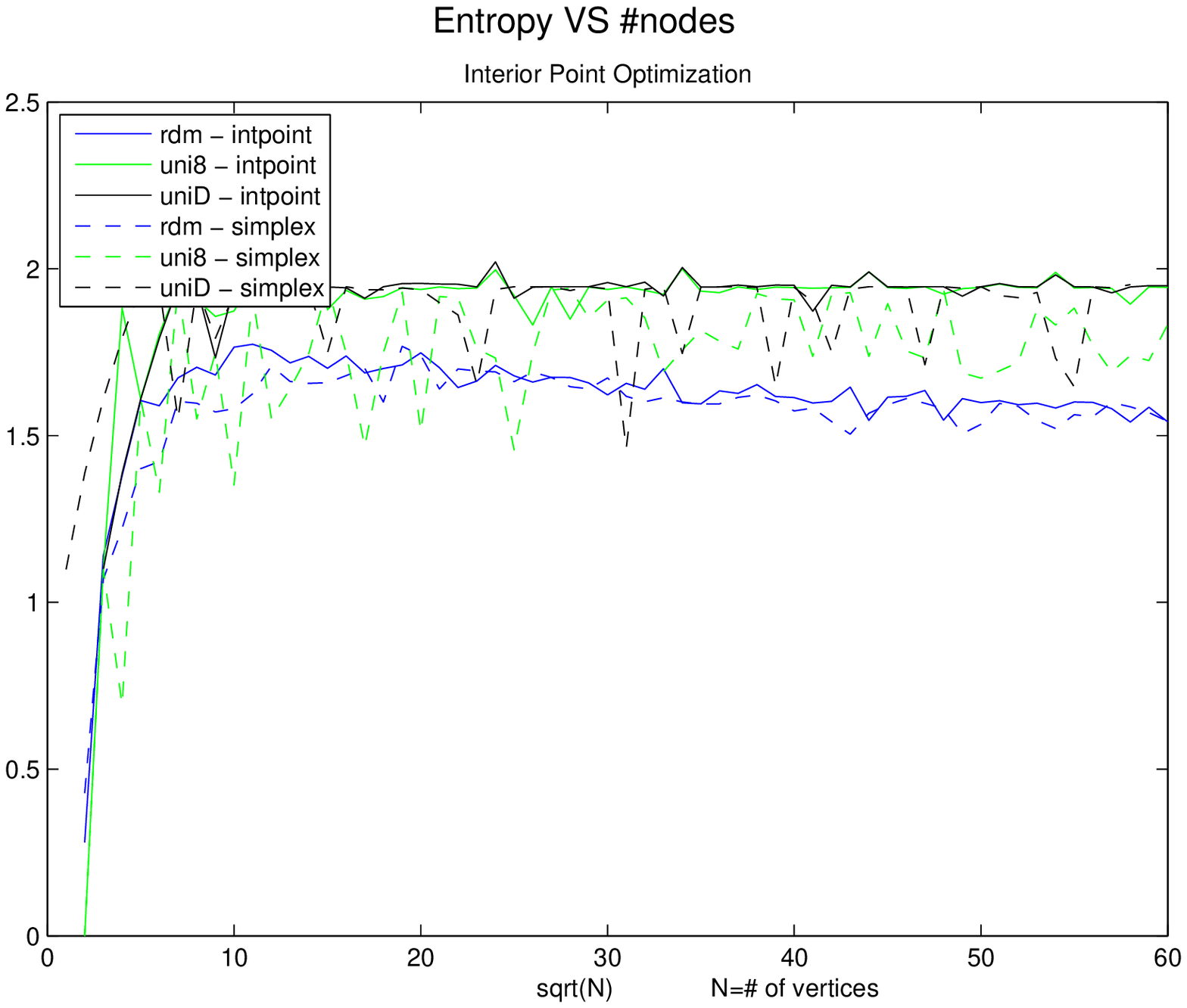}
\label{fig:entropy_vs_nodes_OPTIMIZATION_noE}}\qquad
\subfigure[Entropy  - $\lambda = 10^{-4}$]{\includegraphics[width=.30\textwidth,height=.18\textwidth]{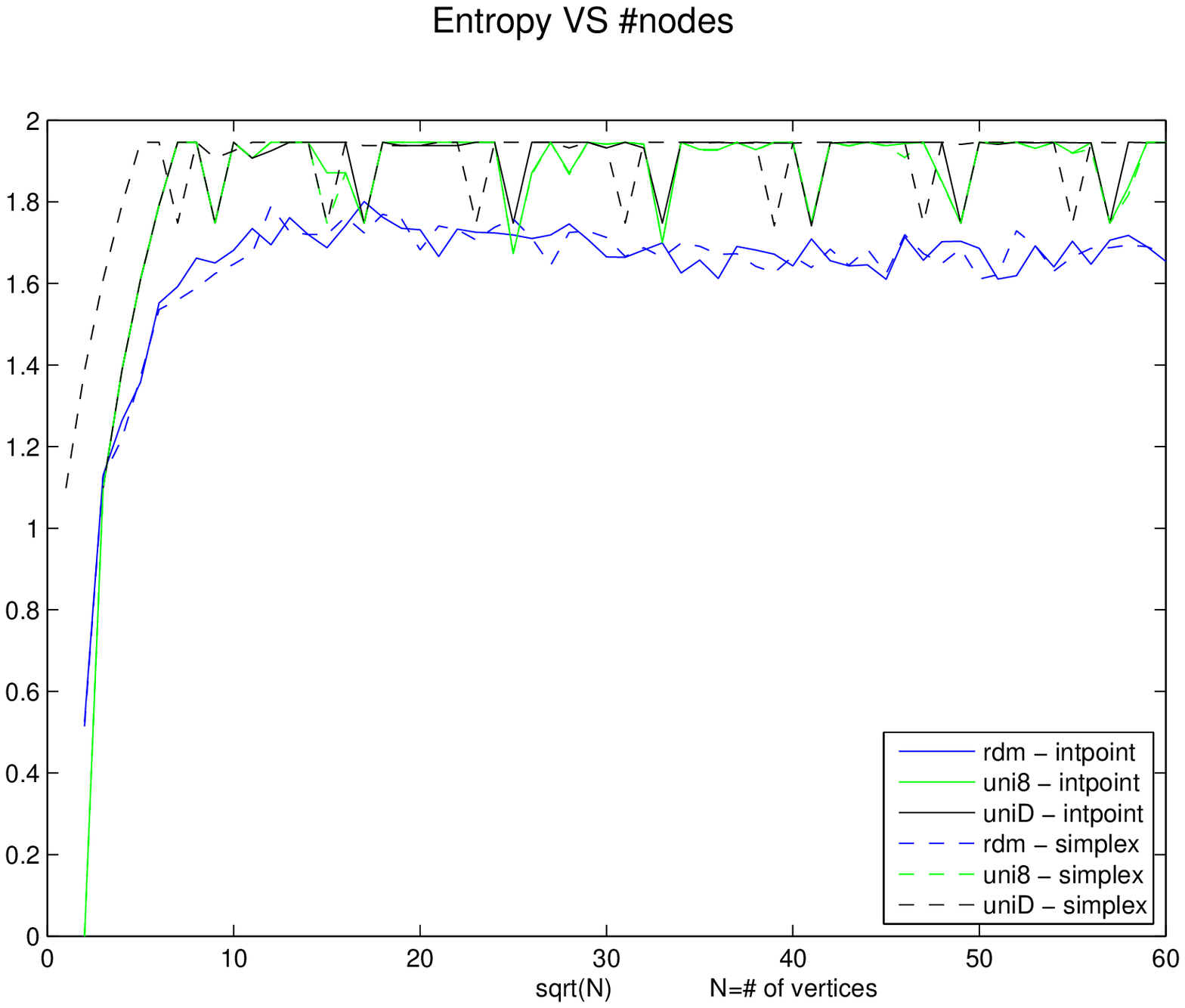}
\label{fig:entropy_vs_nodes_OPTIMIZATION_withE}}\\
\caption{Metrics convergence. The metrics are plotted as function of $\sqrt{N}$. All metrics convergence (except the energy metric) for the three methods for both values of $\lambda$. \texttt{rdm} creates slightly less optimal strategies regarding the entropy for which \unih and \uniD converge to the theoretical value of $\log 7$. \texttt{rdm} also has a slower convergence rate regarding the outcome.}
\label{fig:metrics_vs_nodes_OPTIMIZATION}
\end{figure*}

The results of the sensitivity analysis are represented in Figure \ref{fig:metrics_vs_nodes_OPTIMIZATION}. The metrics defined above are computed for each network construction method, using the simplex or the interior point optimization algorithms, with or without energy optimization, are compared. The energy optimization was introduced to avoid the presence of cycles in the solutions. However, the parameter $\lambda$ needs to remain small compared to $1$, otherwise it would interfere with the strategic optimization goal.
%\begin{itemize}
%\item construction method
%\item size of the network
%\item optimization method
%\end{itemize}
Figure \ref{fig:metrics_vs_nodes_OPTIMIZATION} display the metrics as function of the number of vertices in the network representation of the environment. Comparing them for the different network construction methods and optimization algorithms support the choices to make as to which one are more suited to build an optimal strategy. For the random sampling method, the metrics displayed correspond to the average over ten runs for every network size $N$. 

In Figures \ref{fig:outcome_vs_nodes_OPTIMIZATION_noE},\ref{fig:outcome_vs_nodes_OPTIMIZATION_withE}, the strategic outcome converges for each optimization method and construction choice of the network. Regular networks, such as \unih and \uniD converge faster with or without energy optimization. As soon as there exists a path for each of the seven distinct sections in the environment described above, the solutions are close to optimality. A similar behavior is observed in Figures \ref{fig:entropy_vs_nodes_OPTIMIZATION_noE},\ref{fig:entropy_vs_nodes_OPTIMIZATION_withE}.
In Figures \ref{fig:spreadness_vs_nodes_OPTIMIZATION_noE},\ref{fig:spreadness_vs_nodes_OPTIMIZATION_withE}, two behaviors are observed depending on whether the energy is optimized. On average, more spreading is present when constructing a random network. When optimizing the energy, the regular networks exhibit less spreading, and the spreading converges to a smaller value. Also, the simplex method tends to provide less spread solutions when there is no energy optimization.
In Figures \ref{fig:energy_vs_nodes_OPTIMIZATION_noE},\ref{fig:energy_vs_nodes_OPTIMIZATION_withE}, the random networks are suboptimal in terms of energy repartition. However the energy optimization factor forces the convergence. For regular networks, the energy factor converges quickly. For $\lambda =0$, the space of solutions is larger than for $\lambda= 10^{-4}$. Therefore, it is in the first case that the simplex and interior point methods give the most different results. The spreading and energy metrics tend to favor opposite solutions.

%\nblue{- \textbf{Need for energy optimization} is justified by Figure \ref{fig:energy_vs_nodes_OPTIMIZATION_noE}, where the lack of such an optimization leads to an expected length of travel that grows as $O(N^2)$.}
Figure \ref{fig:energy_vs_nodes_OPTIMIZATION_noE} shows that the lack of such an optimization can lead to an expected length of travel that grows as $O(N^2)$. This property justifies the need for a non-zero energy coefficient $\lambda$.
Comparing the network construction methods, \unih and \uniD offer equivalent results while random sampling is more energy consuming according to Figure \ref{fig:energy_vs_nodes_OPTIMIZATION_withE} results in lesser entropies on the distribution (Figures \ref{fig:entropy_vs_nodes_OPTIMIZATION_noE},\ref{fig:entropy_vs_nodes_OPTIMIZATION_withE}). Note that the max theoretical entropy of $\log (7) \simeq1.94$ is reached by \unih and \uniD.
%
%Regarding the linear optimization algorithms The Simplex algorithm appears to lead to slightly lesser spreading, entropy and outcome especially when no energy optimization is applied. When $\lambda \neq 0$ the algorithms are relatively equivalent because of the significant reduction of the optimal set due to the energy optimization.
%
%\nblue{Sufficient network density (critical size)}
The metrics and methods used converge after a given density of nodes in the network, for a fixed size of the environment. This corresponds to the presence of an edge between each pair of nodes where ambush areas are located.  The limit observed in Figures \ref{fig:outcome_vs_nodes_OPTIMIZATION_noE},\ref{fig:outcome_vs_nodes_OPTIMIZATION_withE} is reached when the network density becomes large enough so that BLUE's distribution can use use a critical number of ambush areas, which is seven here. Hence there exists a sufficient network density for the representation of a two dimensional continuous environment. Interestingly, beyond critical network size, the solutions become similar to the \textit{Hunter and Bird Game} described in \cite{ruckle1979geometric}. There seems to be an analogy between the number of significant paths in the optimal route and the set of homotopy classes that dividing the median would result in. %The theoretical roots of this property will be detailed in future work.	
%
%\nblue{- measure of homotopy spatial difference in comparison with ambush size. ici a cause de la taille des ambush il suffirait que le reseau comprenne 7 chemins (1 par groupe d'ambushs horizontal) pour pouvoir y trouver une strategie optimale}	
	
Overall better results are obtained using the Interior Point algorithm on a uniformly sampled network connected through Delaunay triangulation and with an energy coefficient $\lambda = 10^{-4}$. This technique will be applied to unstructured environments in the remainder of this paper. 
	
%\onecolumn	
%\twocolumn	

%\nblue{Note: there is a equivalently optimal policy for RED that only uses the few "key" ambush areas}
%(see Fig. 
%\ref{fig:min_opt_sol_red}
%)

%\begin{figure}[!ht]
%\center
%\subfigure{\label{fig:min_opt_sol_red_a}\includegraphics[width=0.22\textwidth, height=0.22\textwidth]{uni8_3600nodes_alpha10000_energy0_0000_obs2_intpoint_areaAmbush2_8_voronoi_proba.png}}
%\subfigure{\label{fig:min_opt_sol_red_b}\includegraphics[width=0.22\textwidth, height=0.22\textwidth]{uni8_3600nodes_alpha10000_energy0_0000_obs2_intpoint_areaAmbush2_8_voronoi_proba.png}}
%\caption{Two equivalently optimal solutions for RED. one should have much more ambush sites, with the same game value }
%\label{fig:min_opt_sol_red}
%\end{figure}

%%%%%%%%%%%%%%%%%%%%%%%%%%%%%%%%%%%%%%%%%%%%%%%%%%%%%%%
\section{\label{sec:applications}Applications}

	%%%%%%%%%%%%%%%%%%%%%%%%%%%%%%%%%%%%%%%%
	\subsection{General framework description}

The environment is analyzed for three different purposes. The first objective consists in completing the existing road network with an off-road network to cover both structured and unstructured parts of the environment. The second is to identify relevant geographical areas for ambushes. The third is to compute the local outcome map. A correlation is established at each location between the possible outcome of an ambush and the strategic characteristics of the local environment around this ambush.

\begin{figure}[!ht]
\begin{center}
\begin{tikzpicture}[scale=0.4,transform shape]
 
  % Draw diagram elements
  \path \block {1}{Analytical Environment creation};
  \path (p1.south)+(0.0,-1.8) \data{2}{Road Network};
  \path (p1.south)+(5.0,-1.8) \data{3}{Offroad Environment};
  \path (p3.south)+(0.0,-1.5) \block{4}{Environment discretization};
  \path (p4.south)+(0.0,-1.5) \data{5}{Offroad Network};
  \path (p5.south)+(-10.0,-2.0) \block{6}{Local outcome computation};
  \path (p5.south)+(-2.5,-2.0) \block{7}{Networks merging};
  \path (p6.south)+(-0.0,-1.5) \data{8}{Local outcome map};
  \path (p7.south)+(0.0,-1.5) \data{9}{Complete discrete representation of the environment};
  \path (p8.south)+(5.0,-3.0) \block{10}{Linear Optimization model}; 
    
  \path (p1.north)+(+2.5,2.5) \data{11}{Topological data\\ \cite{ROG:ROG1631}\cite{url:srtm}};
  \path (p1.north)+(-2.5,2.5) \data{12}{Open Street Map data};
  \path (p10.west)+(-3.5,0.0) \data{13}{Departure \& Arrival location};
  
  \path (p10.south)+(-0.0,-1.5) \data{14}{Stochastic routing strategy};
  \path (p6.west)+(-3.5,0.0) \data{15}{Mission type, vehicle type};
  
  % Draw arrows between elements    
  \path [line] (p1.south) -- +(0.0,-0.7) -- +(+5.0,-0.7)
    -- node [above, midway] {} (p3);
  \path [line] (p1.south) -- +(0.0,-0.7)
    -- node [above, midway] {} (p2);
  \path [line] (p1.south) -- +(0.0,-0.7) -- +(-5.0,-0.7)
    -- node [above, midway] {} (p6);
  
  \path [line] (p3.south) -- node [above, midway] {} (p4);
  \path [line] (p4.south) -- node [above, midway] {} (p5);  
  \path [line] (p2.south) -- +(-0.0,-5.0) -- +(2.5,-5.0)
    -- node [above, midway] {} (p7);
  \path [line] (p5.south) -- +(-0.0,-0.88) -- +(-2.5,-0.88)
    -- node [above, midway] {} (p7);
  \path [line] (p7.south) -- node [above, midway] {} (p9);
  \path [line] (p6.south) -- node [above, midway] {} (p8);
  
  \path [line] (p9.west) -- +(-2.5,0.0) -- +(-2.5,2.0)
    -- node [right, midway] {} (p6);
    
  \path [line] (p9.south) -- +(-0.0,-0.66) -- +(-2.5,-0.66)
    -- node [above, midway] {} (p10);
  \path [line] (p8.south) -- +(-0.0,-1.0) -- +(+5.0,-1.0)
    -- node [above, midway] {} (p10);
    
	\path [line] (p13.east) -- node [midway] {} (p10);  
	\path [line] (p11.south) -- +(-0.0,-0.53) -- +(-2.5,-0.53)
    -- node [above, midway] {} (p1);
    \path [line] (p12.south) -- +(-0.0,-0.5) -- +(+2.5,-0.50)
    -- node [above, midway] {} (p1); 
    
    \path [line] (p10.south) -- node [midway] {} (p14); 
    \path [line] (p15.east) -- node [midway] {} (p6);  
   
  \background{p1}{p1}{p1}{p1}{Data processing}
  \background{p6}{p2}{p3}{p9}{Environment analysis}
  \background{p10}{p10}{p10}{p10}{Optmization}
\end{tikzpicture}
\caption{Framework information pipeline.}
\label{fig:pipeline}
\end{center}
\end{figure}

To achieve these objectives, data of various natures is used. The elevation information about the environment is collected through data from the Shuttle Radar Topography Mission \cite{ROG:ROG1631}, which are publicly available \cite{url:srtm}. 
SRTM data is distributed in two precision levels: SRTM1 (for the U.S. and its territories and possessions) with data sampled at one arc-second intervals in latitude and longitude, and SRTM3 (for the entire world) sampled at three arc-seconds. Three arc-second data are generated by three by three averaging of the one arc-second samples. An assessment of the SRTM data precision is given in \cite{Rodriguez2005JPL}. While it is not precise enough for real-time high precision path planning, which is not our present concern, it gives sufficient information for route planning purposes on scales of approximately one to tens of kilometers. 
A typical scenario takes place over 25-30 km in width, with points sampled every hundred meters. The second type of data used in this framework is Open Street Map\cite{Haklay:2008:OUS:1477057.1477249} data. Open Street Map is a global, collaborative effort to create an open map of the world. The data aggregates semantic and usage information about the environment infrastructure such as the type of roads (interstate, primary, secondary, etc), authorized direction (one-way or two-ways), maximum speed, street names or information about the buildings.

The global framework pipeline is presented in Figure \ref{fig:pipeline}.  
%\nblue{describe Figure pipeline}
The different sources of data help create the analytic environment. This feeds into three layers: on the one hand the local outcome map, on the other hand the off-road environment and road information, which merge into the network to be used for the optimization. These three combined lead to the optimization model resulting into the stochastic optimal route. 

As opposed to classical approaches, which require a number of manual operations and fine tuning, the methodology proposed offers an easy straightforward way to find routing strategies. Given the coordinates of the departure and arrival nodes and the type of vehicle, a local outcome map is generated and the optimal stochastic route is obtained. 

	%%%%%%%%%%%%%%%%%%%%%%%%%%%%%%%%%%%%%%%%
	\subsection{Risk factors}

%\nblue{Environmental Risk (risk diagram)}

In order to extend the purely theoretical game theory problem and make it more realistic, interviews of former Army personnel were performed. The focus of these interviews was to identify the factors with the highest influence on the local outcome that could be incorporated into the optimization problem formulation.
Significant attention was devoted to the identification of the different factors influencing the local outcome.

\begin{figure}[!ht]
\begin{center}
\includegraphics[width=.35\textwidth,height=.25\textwidth]{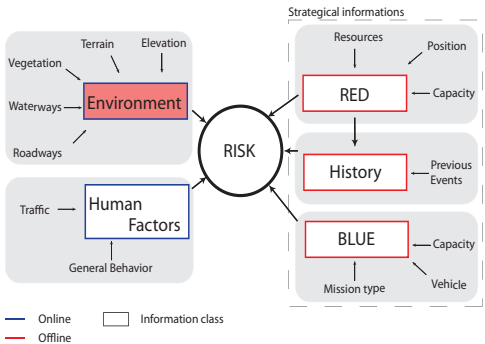}
\caption{Different factors to consider for local outcome computation at a given location.}
\label{fig:risk_diagram}
\end{center}
\end{figure}

The relevant information can be categorized into three different areas represented in Figure \ref{fig:risk_diagram}. First, the information related to the environment, whether described by topological or geospatial data, is decisive. This type of information covers data on elevation, roadways, waterways as well as vegetation and terrain type and quality. Second, the strategical information, such as RED's and BLUE's resources (position, amount, type) as well as previous ambushes, needs to be taken into account. 
Third, the human factors correspond to other intelligent agents present in the geographic region of interest that do not participate in the game but whose behavior may provide additional information. The traffic, the general behavior of a crowd or the amount of people in a given location are examples of such information.
Note that these categories are correlated. For example, for a given type of vehicle, the maximum speed depends on the topology of the environment. A car might be faster than a human on a road but not on steep terrain. Such dependencies are accounted for in the information processing phase.
	
%\nred{need new diagram for risk with 3 categories}

One of the key factors that influence the local outcome of an ambush at a given location is the ability to apprehend a situation as early as possible. Such ability encompasses detecting changes, identifying unusual behavior or setting. This may rely on vision capabilities, in particular sight range and quality. Another key factor consists in having means to react quickly and possibly leave the location. Hence, the player's local maximum speed has a direct impact on the local outcome of an ambush at a given location.

The local outcome map heavily depends on the mission strategical information. The same topological environment, with the same resource allocation between the players, produces a different outcome map depending on the type of mission. Three typical missions are briefly described below as examples. A scouting mission means that BLUE wants to travel from its departure point to its destination point, without being seen at all. The local outcome is then binary : if BLUE is seen, it looses. A transport mission aims at moving goods or assets from A to B. The local outcome is then directly linked to the proportion of the convoy that arrives safely to B. The opposite mission in terms of local outcome is a patrolling mission. The goal is to travel from A to B but a potential encounter would result in a win for BLUE. The local outcome is therefore positive if such an encounter occurs, which is not the case in a transport mission.

	%%%%%%%%%%%%%%%%%%%%%%%%%%%%%%%%%%%%%%%%
	\subsection{Examples}
	
	In the previous section, several factors influencing the local outcome were identified. The remainder of the paper focuses on examples of applications. The local outcome $\alpha$ is now defined as a parameter depending only on the maximum speed, and inversely proportional to it. A slower speed implies a higher local outcome, and reciprocally. The environmental data is used to find the maximum speed at a given position. 
	
%%%%%%%%%%%%%%%%%%%%	
\subsubsection{Road network}\

\begin{figure}
\center
\subfigure[Allowed speed in the city of Monaco.]{\label{fig:osm_speed_profile}\includegraphics[width=0.22\textwidth, height=0.20\textwidth]{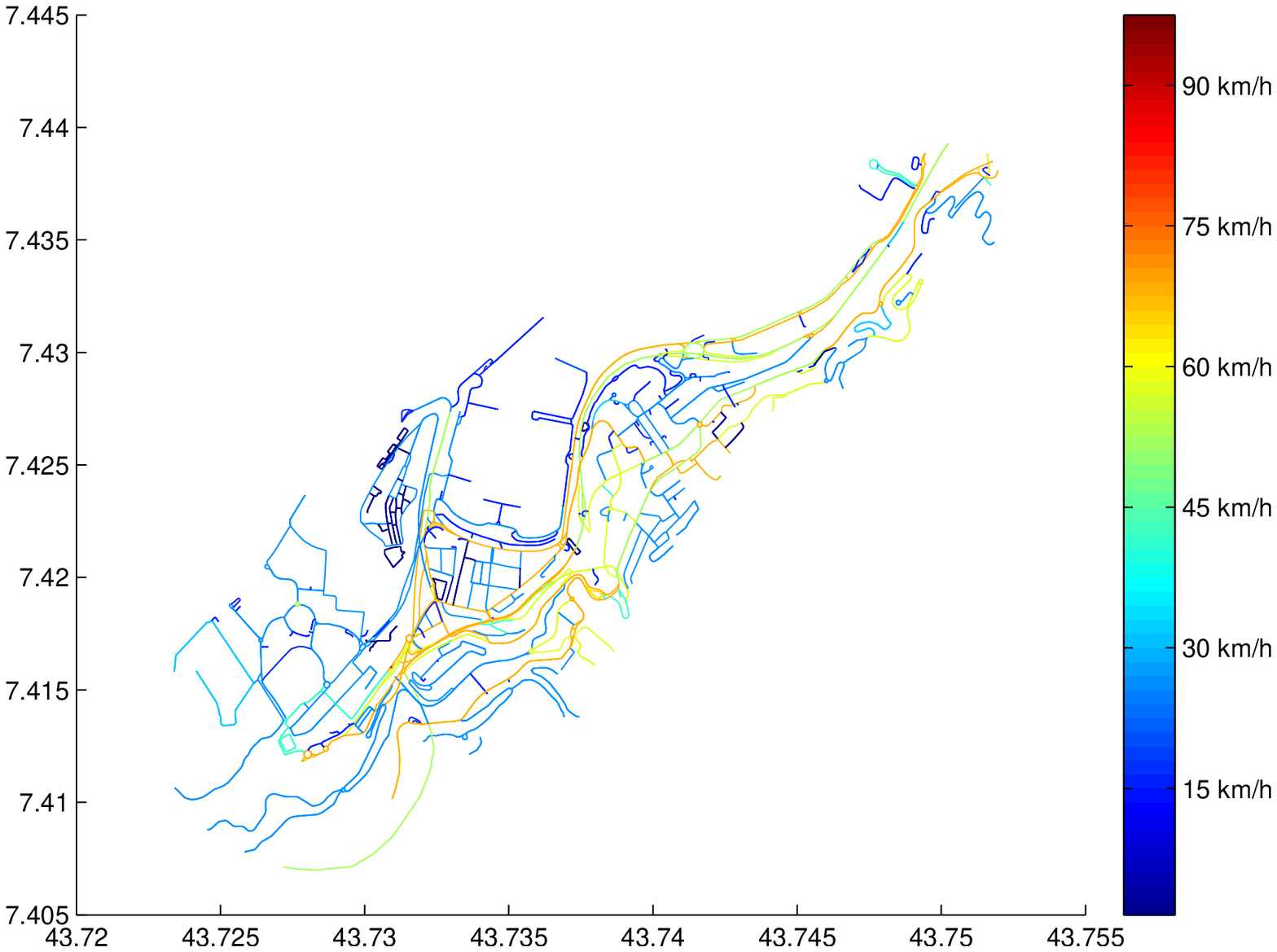}}\qquad
\subfigure[Optimal strategy]{\label{fig:osm_solution}\includegraphics[width=0.22\textwidth, height=0.20\textwidth]{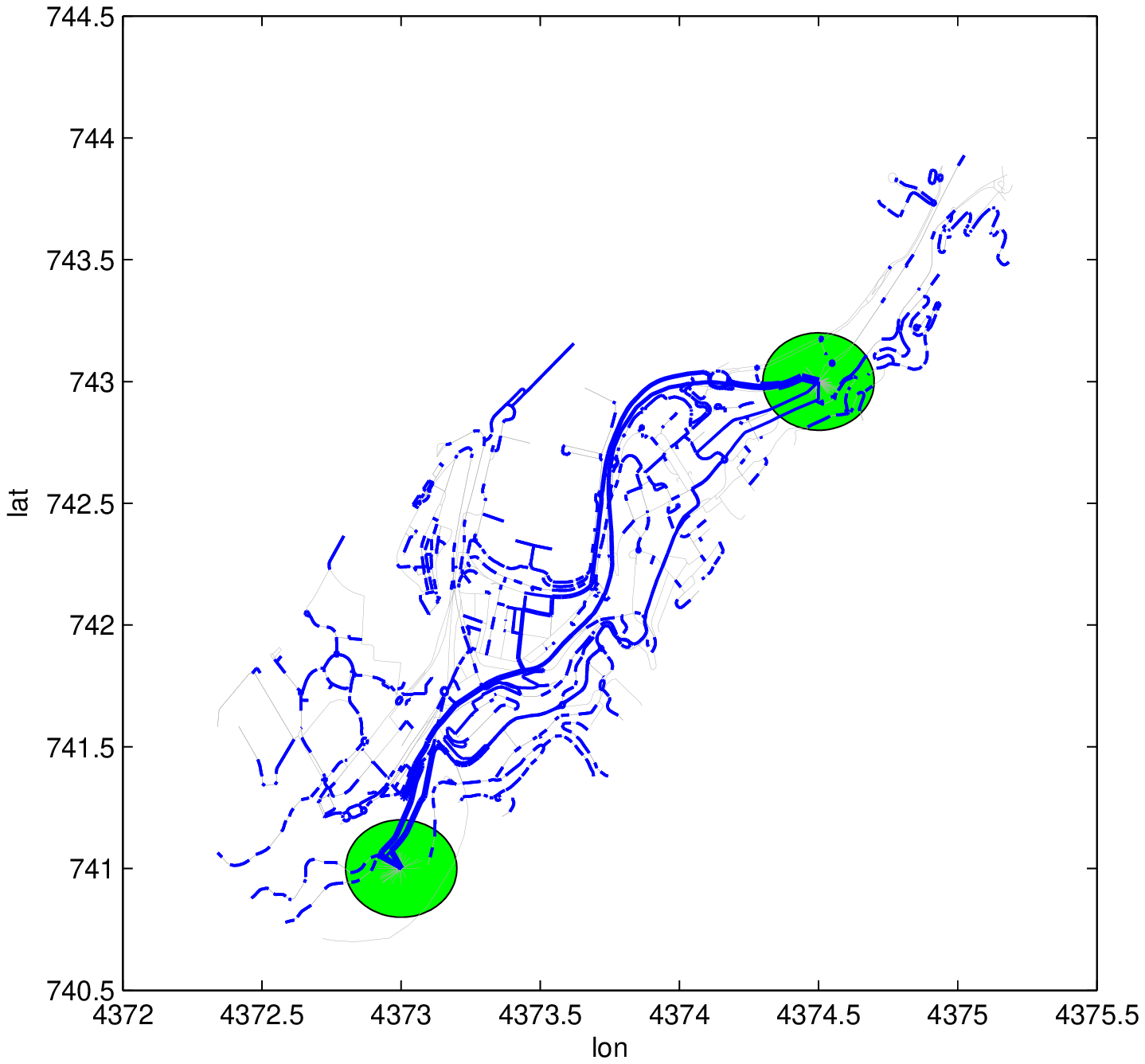}}%
\caption{Example of solution for the road network of the city of Monaco imported from OSM data.}
\label{fig:monaco}
\end{figure}

For the example of the city of Monaco, the portion of unstructured environment is negligible. Therefore the model of ambushes at nodes is used. On road networks the maximum speed is set to be the maximum allowed speed, or the maximum safe driving speed. The local outcome at a given node is proportional to the mean time spent on the edges leading to this node. The speed profile of Monaco is presented in Figure \ref{fig:osm_speed_profile}. The corresponding optimal solution is displayed in Figure \ref{fig:osm_solution}. The departure and arrival nodes are located on opposite sides of the city.

As one might expect, the optimal strategy returned by the framework uses with much higher probability the portions of the network where the speed limit is higher. However, the entire network is used by the routing strategy. 

%%%%%%%%%%%%%%%%%%%%
\subsubsection{Unstructured Environment}\ 

\begin{figure}
\center
\includegraphics[width=0.30\textwidth, height=0.20\textwidth]{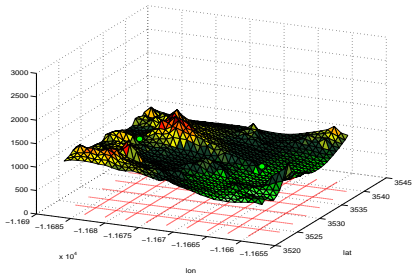}
\caption{3D representation of the environment near Fort Irwin, CA. Imported from STRM data.}
\label{fig:realD_env}
\end{figure}

	On the unstructured part of the environment, the maximum speed is a combined function of the different topological factors and information about BLUE's resources such as its means of locomotion. A three dimensional representation of the environment is displayed in Figure \ref{fig:realD_env}, where lowlands and hills are present.
	
\begin{figure}
\center
\subfigure[Pedestrian]{\label{fig:realD_walk}\includegraphics[width=0.22\textwidth, height=0.20\textwidth]{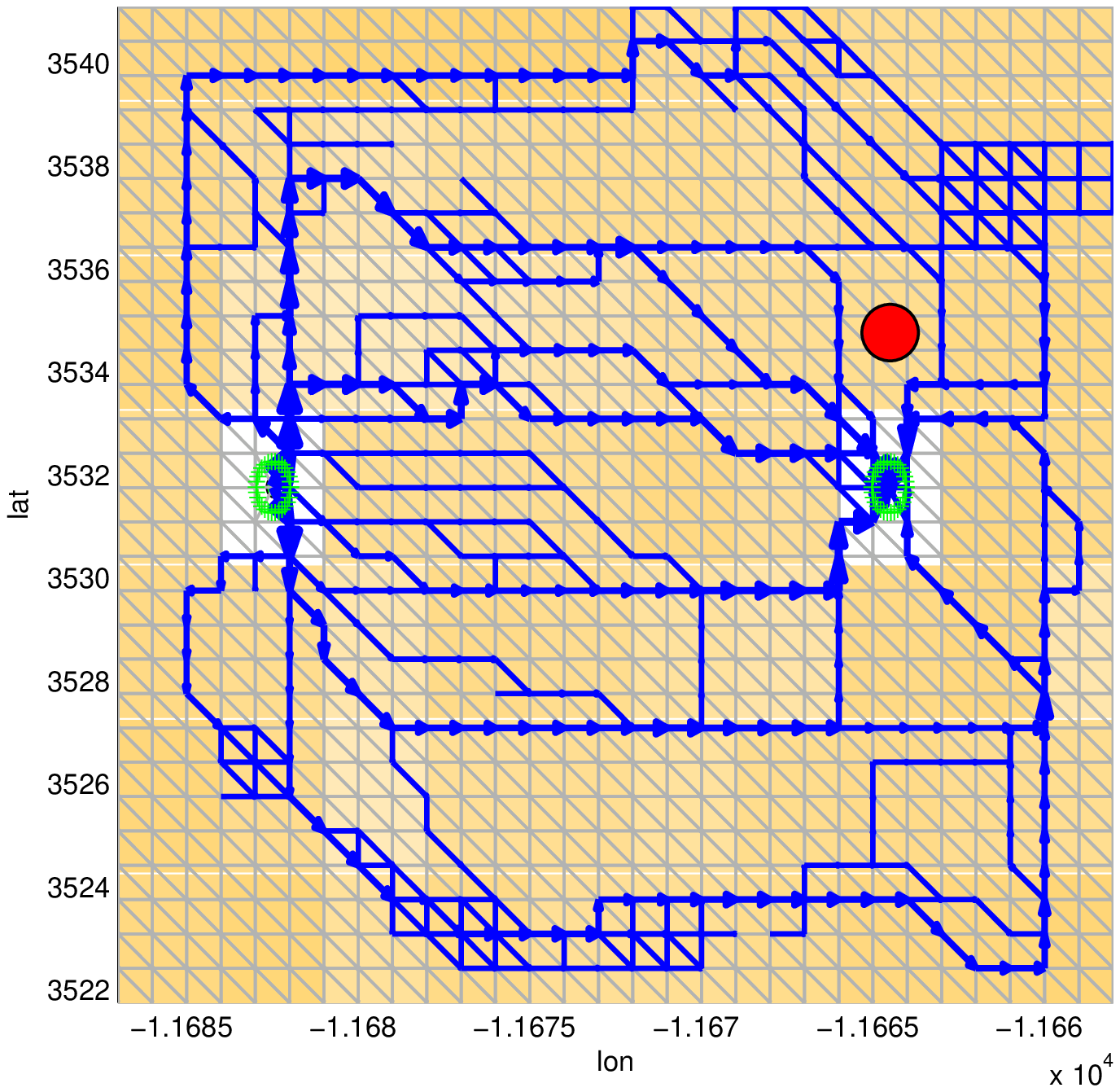}} \qquad
\subfigure[Car]{\label{fig:realD_car}\includegraphics[width=0.22\textwidth, height=0.20\textwidth]{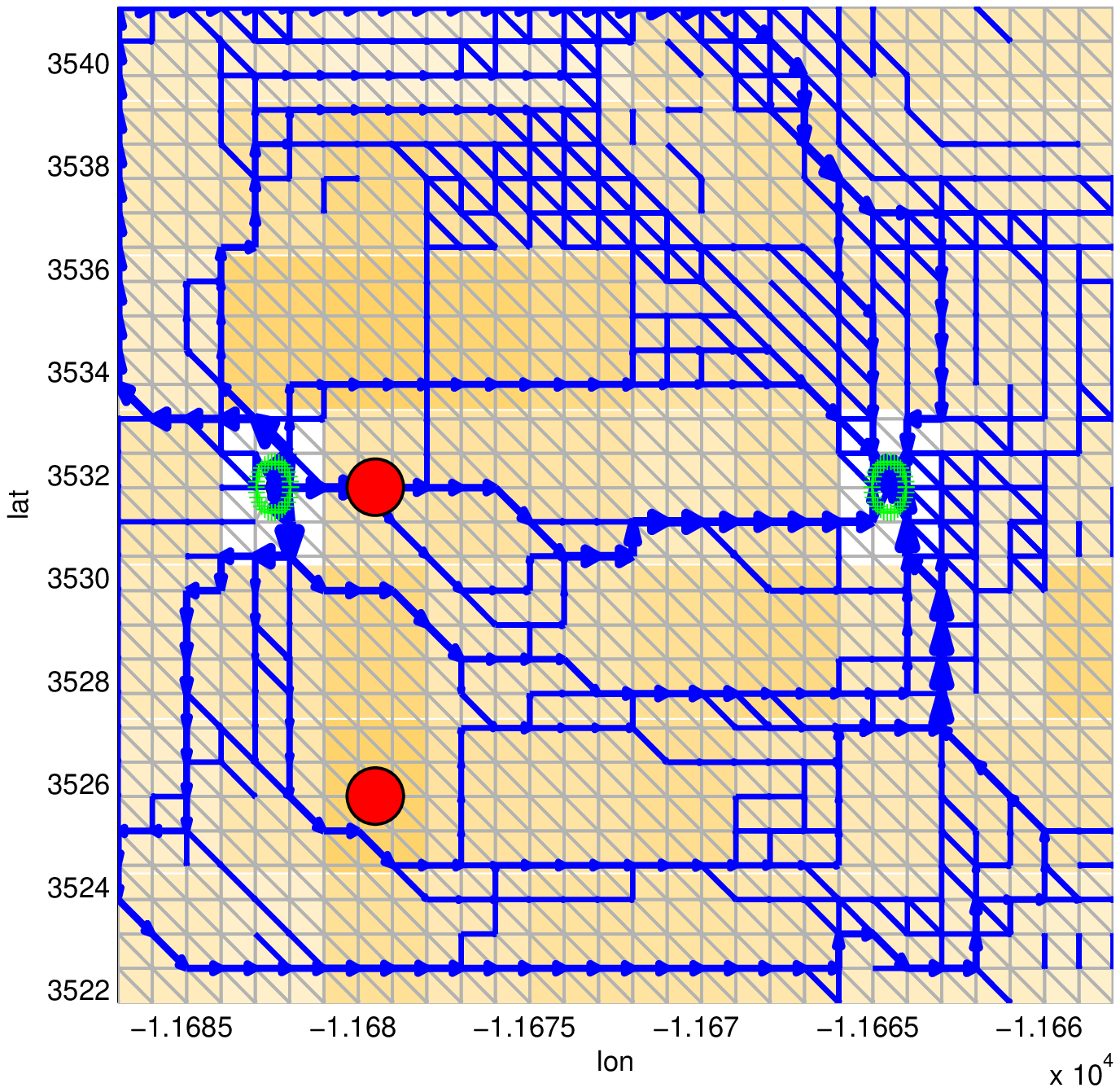}}%
\caption{Optimal routing strategy for a pedestrian and a car near Fort Irwin, CA. The pedestrian is slow and therefore its local outcome is high everywhere. The car's outcome is low in the lowlands (clear color) because it can go much faster. The pedestrian is comparatively less penalized in the hills than the car because its speed is less dependent on the terrain topology. }
\label{realD_compare_car_walk}
\end{figure}

The resulting optimal strategies for either a pedestrian or a car are displayed in Figures \ref{fig:realD_walk} and \ref{fig:realD_car}. The variations of topology in the environment result in local outcome maps that are different for two different means of transportation. The pedestrian being much slower than the car in the lowlands, the local outcome is larger. On the contrary, it is much less penalized by the hills than the car is. Therefore it is more advantageous for the pedestrian to travel through the hills than it is for the car. 

The idea that some resources for BLUE would be more adequate given a specific environment can lead to an interesting expansion to multimodal routing strategies. Given a set of transportation options, a player could change part of its dynamics during its travel in order to reduce the optimal outcome.

Overall the framework developed is efficient. Since it requires few inputs from the user, it can be easily used to compare different routing strategies on various environments. The large amount of SRTM and OSM data available makes it applicable to almost anywhere in the world.

%%%%%%%%%%%%%%%%%%%%%%%%%%%%%%%%%%%%%%%%%%%%%%%%%%%%%%
\section{Future Work}

%\nblue{Mathematical model for analytical resolution. Homotopies}
From the theoretical standpoint, several aspects could be further investigated. The convergence observed for the routes among all methods suggests the existence of a limit distribution. The authors would like to propose a continuous model of the game at hand instead of the dense discretization developed in this paper. This would allow to identify the limit distribution at once and maybe draw a parallel between the distinct paths among routes and homotopy classes.

%\nblue{merging structured and unstructured}
%\nblue{Lower scale risk: Scene analysis}
%\nblue{better risk}
A number of applications have been envisioned. The only factor influencing the local outcome is currently the maximum speed at a given location. The fidelity of the model could be improved by taking into account more risk factors in more detail. 
The purpose of this framework is to provide a hands-on tool for route prediction. However the merging of structured and unstructured environments still requires manual tuning. This merging process will be automated more thoroughly in the upcoming work.
Finally, it might be interesting to increase the precision of the environment description by one or two orders of magnitude. Doing so requires determining how to accurately forecast the maximum safe speed at a given location. The control of the vehicle hence becomes part of the optimization problem.

%%%%%%%%%%%%%%%%%%%%%%%%%%%%%%%%%%%%%%%%%%%%%%%%%%%%%%
\section{Conclusion}
The work developed build on a previous research effort in which the environment was structured and no study of the local outcome of an ambush inference was performed. 
The present research defines the opponents reach as an important parameter of the environment representation. The local outcome becomes more meaningful because of its geographical significance. 
The possible losses for the system at a given area in case of ambush are computed through the identification of the different factors influencing this outcome. These factors are related to the environment and to the agents' resources.
Comparing different network construction methods and different linear optimization algorithms, efficient techniques were identified to elaborate ambush avoidance strategies on unstructured environments. 
Results highlight the existence of a sufficient network density to represent these environments. The correlation between the reach of the ambushing player and the optimal outcome of the game is illustrated and a "representative distribution" emerges. 
Finally a comprehensive framework is elaborated. It takes as input topological data and, given a pair of points and a vehicle type, provides an optimal stochastic routing strategy between these two points. Examples of application are tested for structured and unstructured environments.

% use section* for acknowledgement
\section*{Acknowledgement}
% optional entry into table of contents (if used)
% \addcontentsline{toc}{section}{Acknowledgment}
This work was supported by the Army Research Office under MURI Award
W911NF-11-1-0046.

%% Use plainnat to work nicely with natbib. 

\bibliographystyle{plainnat}
\bibliography{references}

\end{document}